\documentclass[]{interact}

\usepackage{epstopdf}%
\usepackage[caption=false]{subfig}%

\usepackage[numbers,sort&compress]{natbib}%
\bibpunct[, ]{[}{]}{,}{n}{,}{,}%
\renewcommand\bibfont{\fontsize{10}{12}\selectfont}%
\makeatletter%
\def\NAT@def@citea{\def\@citea{\NAT@separator}}%
\makeatother%

\theoremstyle{plain}%

\theoremstyle{definition}

\theoremstyle{remark}

\begin{document}

\articletype{Full Paper}%

\title{Variable-Speed Teaching--Playback as Real-World Data Augmentation for Imitation Learning}

\author{
  \name{
    Nozomu Masuya\textsuperscript{a}\thanks{CONTACT Nozomu Masuya. Email: masuya.nozomu.sm@alumni.tsukuba.ac.jp}, 
    Hiroshi Sato\textsuperscript{a}, 
    Koki Yamane\textsuperscript{b}, 
    Takuya Kusume\textsuperscript{a}, 
    Sho Sakaino\textsuperscript{c}, and 
    Toshiaki Tsuji\textsuperscript{d}
  }
  \affil{
    \textsuperscript{a}Master's Program in Intelligent and Mechanical Interaction Systems, Degree Programs in Systems and Information and Engineering, Graduate School of Science and Technology, University of Tsukuba, Japan; 
    \textsuperscript{b}Doctoral Program in Intelligent and Mechanical Interaction Systems, Degree Programs in Systems and Information and Engineering, Graduate School of Science and Technology, University of Tsukuba, Japan; 
    \textsuperscript{c}Department of Intelligent Interaction Technologies, Institute of Systems and Information Engineering, University of Tsukuba, Japan;
    \textsuperscript{d}Graduate School of Science and Engineering, Saitama University, Saitama, Japan.
  }
}

\maketitle

\begin{abstract}
  Because imitation learning relies on human demonstrations in hard-to-simulate settings, the inclusion of force control in this method has resulted in a shortage of training data, even with a simple change in speed.
  Although the field of data augmentation has addressed the lack of data, conventional methods of data augmentation for robot manipulation are limited to simulation-based methods or downsampling for position control.
  This paper proposes a novel method of data augmentation that is applicable to force control and preserves the advantages of real-world datasets.
  We applied teaching--playback at variable speeds as real-world data augmentation to increase both the quantity and quality of environmental reactions at variable speeds.
  An experiment was conducted on bilateral control-based imitation learning using a method of imitation learning equipped with position--force control.
  We evaluated the effect of real-world data augmentation on two tasks, pick-and-place and wiping, at variable speeds, each from two human demonstrations at fixed speed. 
  The results showed a maximum 55\% increase in success rate from a simple change in speed of real-world reactions and improved accuracy along the duration/frequency command by gathering environmental reactions at variable speeds.
\end{abstract}


\section{Introduction}

  Particularly in societies with a declining workforce, robots are expected to replace a wider range of human labor, such as handling irregularly shaped objects or conducting multiple household tasks. 
  As the precise modeling of contact or irregularly shaped objects is infeasible, model-free methods are required to expand the workplace for robots. 
  An example of a model-free method is reinforcement learning~\cite{levine2018}. 
  Although this method is completely model-free, it requires numerous attempts, which are difficult to achieve.
  
  Imitation learning, such as behavioral cloning or learning from demonstrations, balances the requirements of prior knowledge and iterations on real robots with the help of experts. 
  In particular, imitation learning based on teleoperation, such as ALOHA~\cite{aloha} and bilateral control-based imitation learning~\cite{adachi2018, akagawaJIA, yamaneral, biact, ilbit} has succeeded in adapting to the environment and generating motion in real time. 
  In these methods, first, teaching is performed through teleoperation by humans, and then motion is reproduced by the robot  imitating human commands. 
  This enables the collection of commands for robots in addition to their responses, making efficient use of human skills in controlling robots based on predictions.
  
  In conventional methods of imitation learning with force control, learning motion along variable-speed commands ~\cite{sakaino2022, CRANEX7params} requires human demonstrations at various speeds in the initial phase of teaching. 
  The application of self-supervised learning~\cite{CRANEX7params}, which utilizes the generated motion as additional training data, has succeeded in increasing the interpolating speed and accuracy. 
  However, this approach still struggles to reproduce faster motion than that originally trained. 

  Because the safe teleoperation of a robot requires speed limitations, this has limited the speed of motion to that of teleoperation to a smaller range than human motion itself. 
  Meanwhile, position--force-controlled teaching--playback, called the motion-copying system~\cite{yokokura2008b}, has succeeded in reproducing motion at variable speeds, including at fast-forward speeds.
  Fast-forward data collection for one task at a fixed speed has been utilized~\cite{sakainosamcon}, but the effect of fast-forward remains unclear, as is the feasibility of teaching tasks at variable speeds.

  The contribution of this paper is the proposal of a method to reduce human effort in teaching imitation learning based on force control through the validation of real-world data augmentation at variable speeds in bilateral control-based imitation learning.
  This study verified real-world spatio-temporal data augmentation using actual robots and the environment and verified the validity of the proposed method through experiments involving two tasks with two human demonstrations.

\section{Related Work}
  \subsection{Imitation Learning for Robot Manipulation}

    As technologies in machine learning, particularly neural networks (NNs), have gained popularity, the field of imitation learning~\cite{lee2015, yang2017, osa2018, hussein2018, fang2019}, including behavioral cloning and learning from demonstration, has attracted interest owing to its high efficiency and versatility. 
    Imitation learning aims to transfer skills from experts to robots through machine learning. 
    In robot manipulation, imitation learning is utilized to extract human skills in handling unknown objects or phenomena, such as handling food or dealing with friction. 

    Similarly, in other fields related to machine learning, the collection of expert data in an appropriate format, as well as algorithms for generating or selecting trajectories, is critical to the success of imitation.
    Data collection methods in imitation learning are largely divided into indirect and direct teaching methods~\cite{fang2019}.
    Indirect teaching collects human demonstrations through the human body, whereas demonstrations in direct teaching are performed by the robot as imitation is expected.
    Although indirect teaching is safe and low-cost, the difference in embodiment between humans and robots makes imitation difficult.

    Direct teaching methods include kinesthetic teaching~\cite{ghalamzan2018}, virtual-reality based teleoperation~\cite{yang2017, zhang2018, si2021}, and leader--follower teleoperation~\cite{adachi2018, aloha}.
    In contrast to kinesthetic teaching, which directly involves the single robot expected to imitate it, virtual-reality- and leader--follower-based teleoperation methods enable the acquisition of commands and responses separately.
    Because robots can have delays, particularly at high frequencies, the separation of commands and responses is crucial to achieving human-like quick motion.
    Therefore, teleoperation-based imitation learning is one of the most promising methods for acquiring human manipulation skills.

  \subsection{Data Augmentation for Robot Manipulation}

    In machine learning, data augmentation is a method for reconstructing diverse real-world datasets from a limited amount and variety of data.
    For visual data, rotation, distortion, masking, mixing, and adding noise are among the well-known examples, in addition to synthesizing from games or simulations~\cite{MUMUNI2022, MAHARANA2022}, or additional data is generated with an NN, such as a generative adversarial network (GAN)~\cite{Bissoto2021}. 

    In robot manipulation, the subjects of data augmentation are widened to include mechanical data such as position and force, in addition to visual or other sensory data.
    Generative approaches, including simulations, are options with maximal diversity when modeling is possible.
    Mitrano \textit{et al}.~\cite{mitrano2022data} proposed a data augmentation based on simulations and optimization. While this approach provides an increased diversity of scenarios preserving the state of contact, it is limited to known environments and objects, owing to the assumptions that make simulation possible.
    Yu \textit{et al. }'s ROSIE~\cite{rosie} can randomize the properties of items in visual data using a text-guided image generation model. Although this method has successfully augmented the appearance of objects, its application to motion and force requires a large dataset from robots, which is currently unavailable.

    In contrast, downsampling and rearrangement are among the easy-to-implement options, as robots are controlled at a high frequency for precision and stability, enabling the acquisition of motion data at higher frequencies than that of cameras or NN models. 
    Rahmatizadeh \textit{et al}.~\cite{rahmatizadeh2018} implemented the downsampling and rearrangement of trajectory data from 33 to 4~Hz in addition to synthesizing additional trajectory data between demonstrations. This method was implemented for bilateral control-based imitation learning~\cite{adachi2018, CRANEX7params, yamaneral, biact}, generating training data from 25 to 100~Hz from mechanical data of up to 1~kHz.
    Kobayashi \textit{et al}.~\cite{kobayashi2024} proposed an efficient multi-modal combination of downsampled mechanical and low-frequency visual data in bilateral control-based imitation learning, minimizing the time difference between the visual and mechanical data.
    A change in speed can also be accomplished by downsampling for limited mechanical data; e.g., Yamamoto \textit{et al}.~\cite{yamamoto2023} augmented the speed and phase of visual data and joint angles through downsampling. While this method enables conveyor picking at variable speeds, it is limited to position control, leaving room for force control. 
    
    Sakaino \textit{et al}.'s fast-forward data collection~\cite{sakainosamcon} included the downsampling of human commands and a real-world collection of environmental reactions, including force, through teaching--playback. This method enables the collection of hard-to-simulate force reactions with minimal human effort; however, its effect and feasibility for variable-speed tasks remain unclear.

\section{Background}
  \subsection{Four-channel Bilateral Control}
\begin{figure}[tbhp]
      \centering
      \includegraphics[scale=0.3, bb = 0 0 737 301]{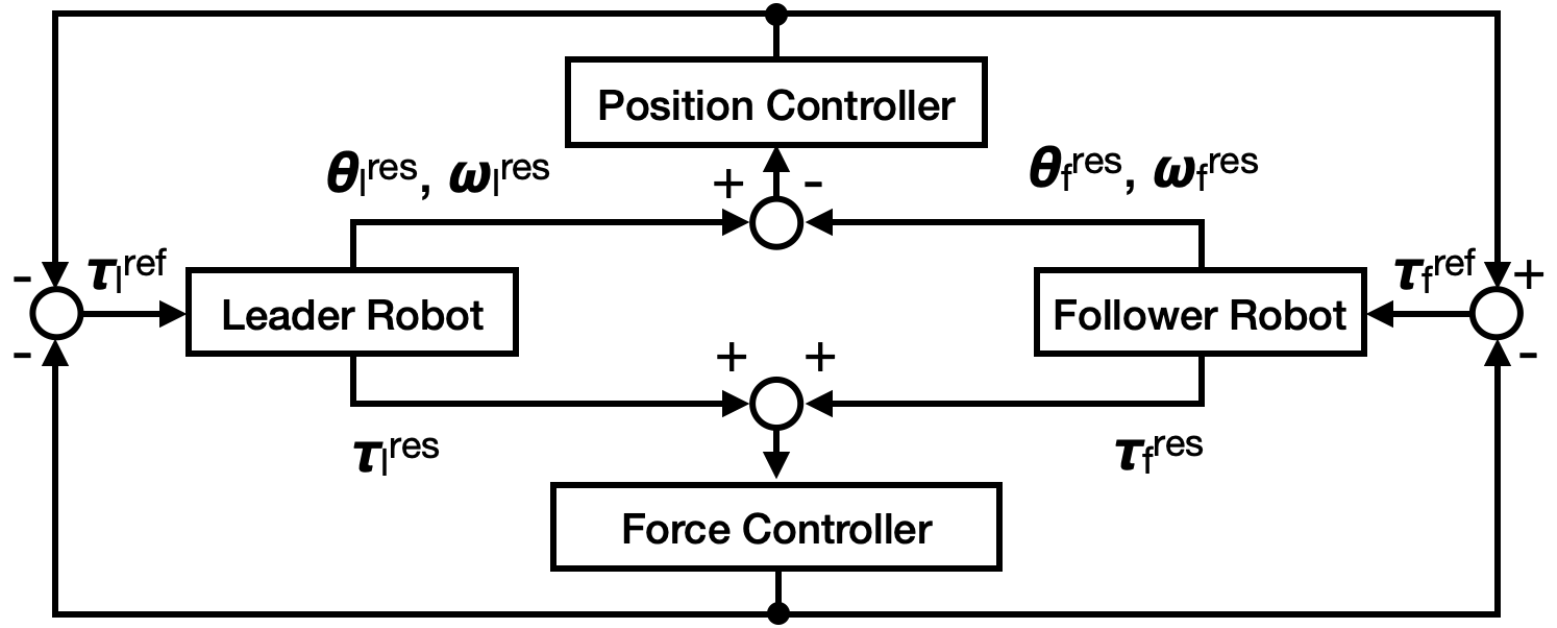}
      \caption{Block diagram of four-channel bilateral control.}
      \label{fig:4ch}
    \end{figure}

    The bilateral teleoperation of robots has long been a field of research for applications in extreme environments, hazardous tasks, or minimally invasive surgeries~\cite{Hokayem2006}.
    Four-channel bilateral control~\cite{tavakoli2007, sakaino2011} is a method of bilateral teleoperation that reconstructs actions and reactions between a leader robot for the operator and follower robot performing the task, with symmetrical feedback of both position and force.
    In four-channel bilateral control, the leader and follower robots are controlled to satisfy the following equations: 
    \begin{equation}
        \bm{\theta}^\mathrm{res}_f - \bm{\theta}^\mathrm{res}_l = \bm {0} 
    \end{equation}
    and
    \begin{equation}
        \bm{\tau}^\mathrm{res}_f + \bm{\tau}^\mathrm{res}_l = \bm {0}, 
    \end{equation}
    where $\bm{\theta}$ is the vector of the joint angle, $\bm{\tau}$ is that of the joint torque, and the subscripts $l$ and $f$ represent the leader and follower, respectively. 
    A block diagram of four-channel bilateral control is shown in figure~\ref{fig:4ch}. 
    The superscripts $res$ and $ref$ denote the response and reference, respectively.
    In four-channel bilateral control, the scaling of the position and force between the leader and follower is available through oblique coordinate control~\cite{sakaino2011}. 
    However, scaling the length of time within bilateral control clearly collides with the law of cause and effect, rendering it infeasible.

  \subsection{Motion-Copying System}
    The motion-copying system~\cite{yokokura2008b, yokokura2009, Igarashi2015, fujisaki2023}, a teaching--playback method with position--force control, is derived from the substitution of the leader robot in four-channel bilateral control with motion data previously obtained through four-channel bilateral control~\cite{yokokura2008b, yokokura2009, Igarashi2015} or an equivalent source of position and force data~\cite{fujisaki2023}. 
    
    This method enables the replay of skilled human motion, obtained in four-channel bilateral control, without continuous teleoperation. 
    In addition to simple replay, motion can be slowed or fast-forwarded~\cite{yokokura2008b, yokokura2009} by altering the playback speed in addition to spatial scaling~\cite{Igarashi2015}.

\begin{figure}[tbhp]
      \centering
      \includegraphics[scale=0.3, bb = 0 0 644 293]{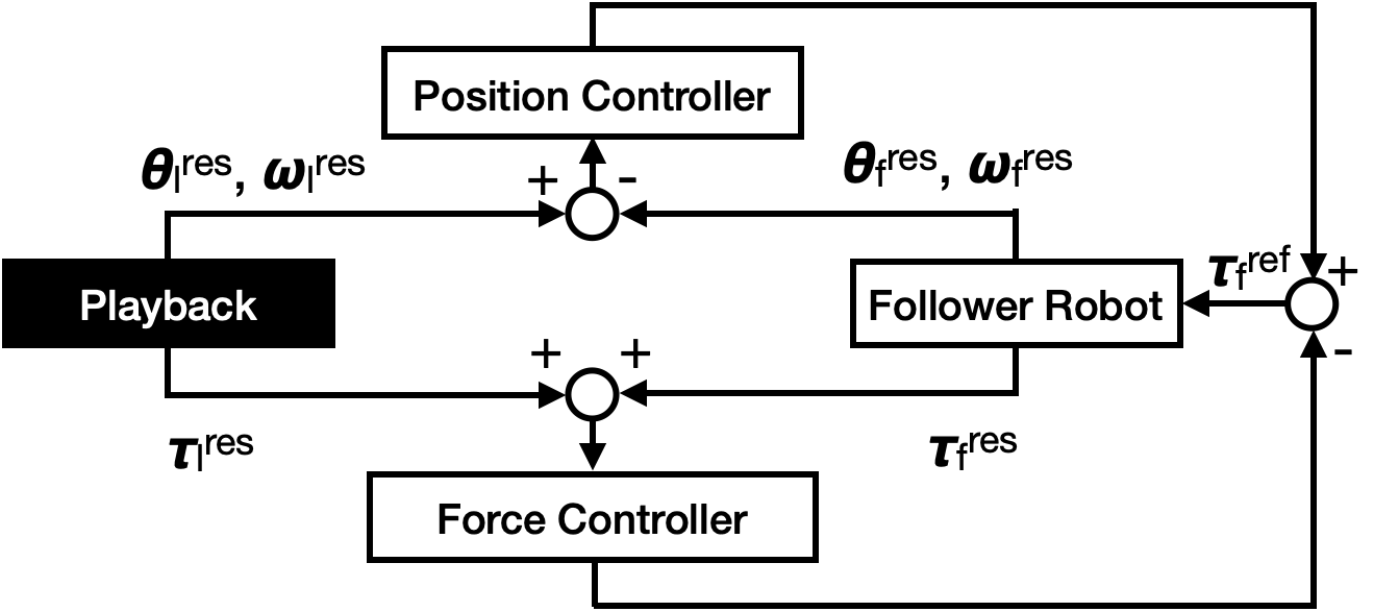}
      \caption{Block diagram of the motion-copying system, teaching--playback with force control.}
      \label{fig:mocopy}
    \end{figure}

    A block diagram of the motion-copying system during the playback phase is shown in figure~\ref{fig:mocopy}. The definitions of the variables, superscripts, and subscripts are the same as those for the four-channel bilateral control.
    Clearly, playback using this method excludes feedback from the environment to the command values. 

  \subsection{Bilateral Control-Based Imitation Learning}
\begin{figure}[tbhp]
      \centering
      \includegraphics[scale=0.3, bb =  0 0 644 293]{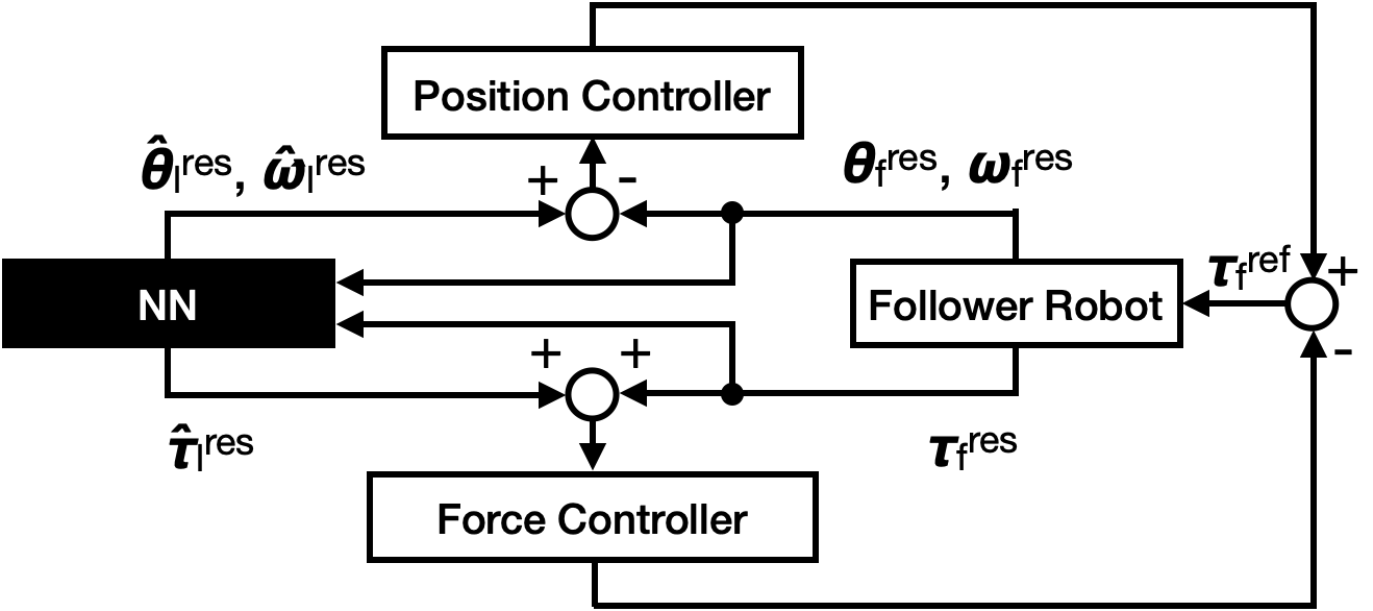}
      \caption{Block diagram of bilateral control-based imitation learning.}
      \label{fig:moho}
    \end{figure}

    Bilateral control-based imitation learning~\cite{adachi2018, sakaino2022, akagawaJIA, yamaneral} is a method of teleoperation-based imitation learning with position--force control. 
    This method is derived from the substitution of the leader robot of the four-channel bilateral control with an NN model trained to reproduce human commands from the state of the follower robot.
    The use of force control and the preservation of the controller enables a robot to perform nonprehensile manipulation~\cite{CRANEX7params}, grasp unknown objects without crushing~\cite{yamaneral}, and perform other contact-rich tasks at the same speed as humans.
    To train the NN model, this method utilizes four-channel bilateral control to obtain the command values of the position and force separately from the response of the follower robot, which makes contact with the environment.
    Figure ~\ref{fig:moho} depicts the block diagram of bilateral control-based imitation learning. The definitions of $\bm{\theta}$, $\bm{\omega}$, $\bm{\tau}$, superscripts, and subscripts are the same as those of the four-channel bilateral control and motion-copying system, and the circumflexes~(~$\hat{}$~) denote the estimates by the NN model.
    The configuration of bilateral control-based imitation learning differs from that of teaching--playback (figure~\ref{fig:mocopy}) in the feedback from the follower robot to the NN model. 

\section{Method: Teaching--Playback at Variable Speeds as Real-World Data Augmentation}

  We utilized real-world teaching playback to obtain many hard-to-simulate environmental reactions.
  Real-world data augmentation was conducted for three main reasons.
  First, because the robot system and its surrounding environment contains nonlinearity, variations in speed invoke nonlinear hard-to-simulate changes in the reaction. 
  Second, as imitation learning utilizes environmental reactions as input, room remains for improvement of accuracy through learning the invariance of commands with a small change in reaction.
  Third, imitation learning in contact-rich tasks with force control requires a dataset of contact forces, whose collection must occur in a real environment.
  
  In this study, bilateral control-based imitation learning was selected as the method for imitation learning to enable force control, and a motion-copying system was used for data augmentation. 
  Linear interpolation was utilized to alter the speed of the motion data, and velocity data, both command and response, were multiplied through the speed ratio. 
  In addition, to address the increased probability of failure owing to the absence of feedback, the motion was repeated until a certain number of successful playbacks were reached according to the tasks, and only successful data were used to train the NN model. 

\begin{figure}[tbhp]
    \centering
    \includegraphics[scale=0.3, bb =  0 0 996 564]{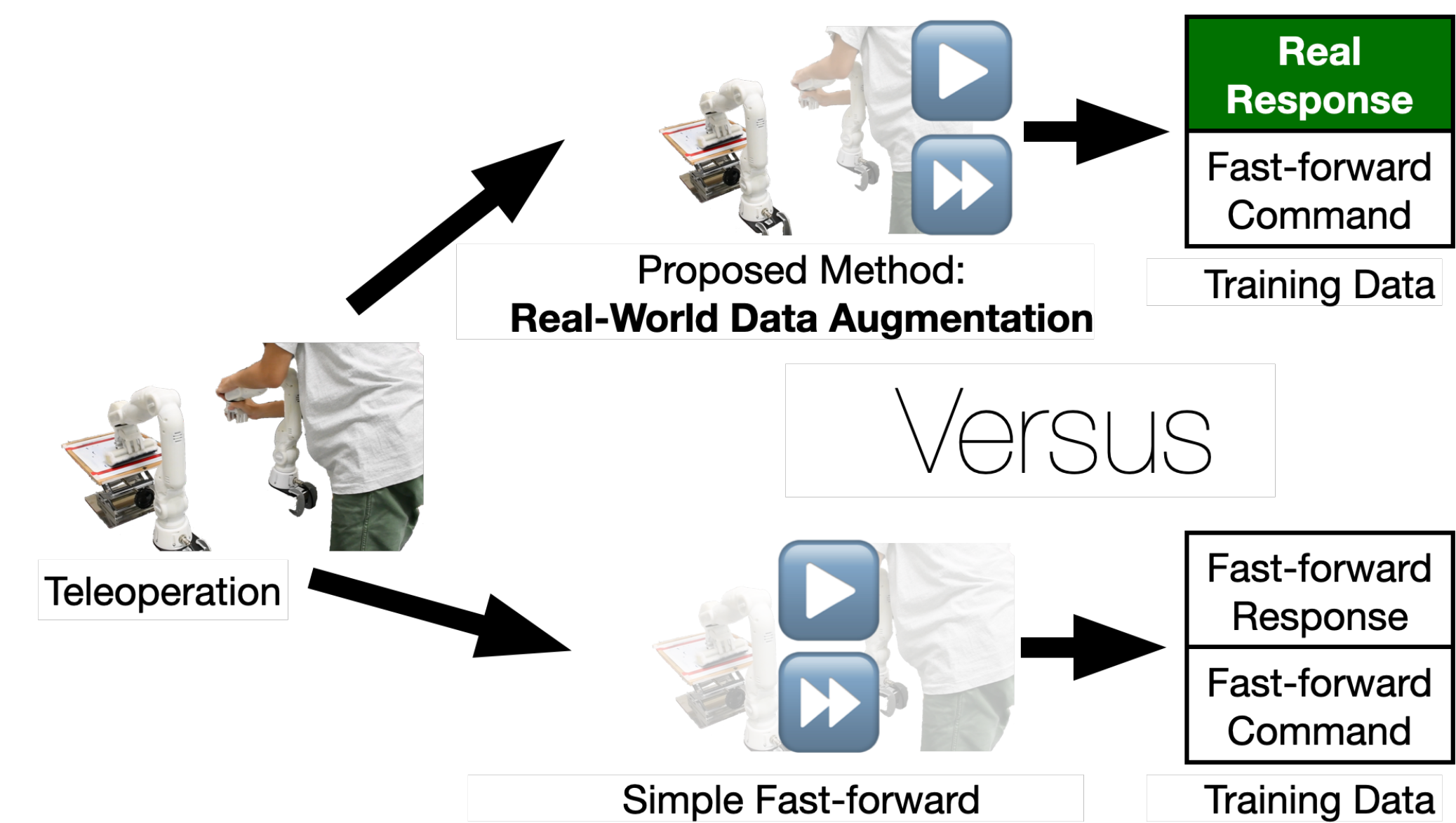}
    \caption{Difference between the proposed method and a simple fast-forward in data.}
    \label{fig:proposal}
  \end{figure}

  The experimental flowchart before training the NN models is shown in figure~\ref{fig:proposal}. 
  Simple duplication and speed adjustments were performed to evaluate the effect of real-world data augmentation. 
  Both methods shared the same command values, but the corresponding response values were different. 
  
\section{Experiment and Evaluation}
  \subsection{Setup of Robots}
    \subsubsection{Manipulator}
\begin{figure}[tbhp]
          \centering
          \resizebox*{5cm}{!}{\includegraphics[scale=0.005, bb = 0 0 3235 3238]{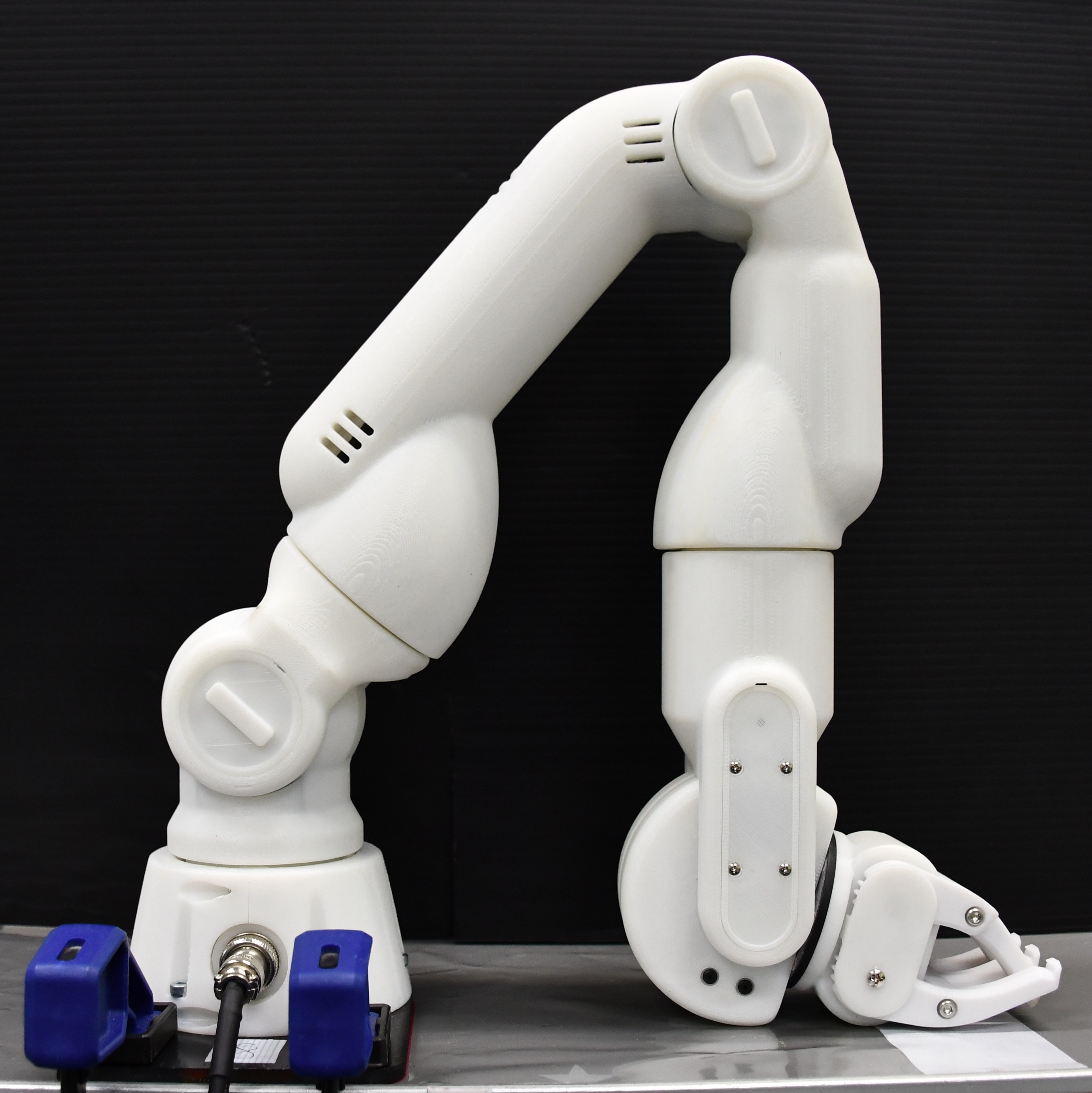}}
          \caption{CRANE-X7 with a cross-structured hand.}
          \label{fig:CRANEX7}
      \end{figure}

      Figure~\ref{fig:CRANEX7} depicts the manipulators used in this study. 
      They are called CRANE-X7 and are manufactured by RT Corporation, Tokyo, Japan.
      The CRANE-X7 consists of an arm with seven degrees of freedom and a rigid end effector with one degree of freedom, each equipped with rotary encoders capable of current control. 
      The end effectors were replaced with a cross-structured hand to enable the grasping of small objects~\cite{yamaneral}. 
      In addition, the third joint was mechanically fixed to eliminate redundant degrees of freedom.
      We used two CRANE-X7 robots in the initial teaching process using bilateral control and one in teaching--playback and evaluation. 

    \subsubsection{Controller for Each Manipulator}
\begin{figure}[tbhp]
          \centering
          \includegraphics[scale=0.25, bb = 0 0 890 318]{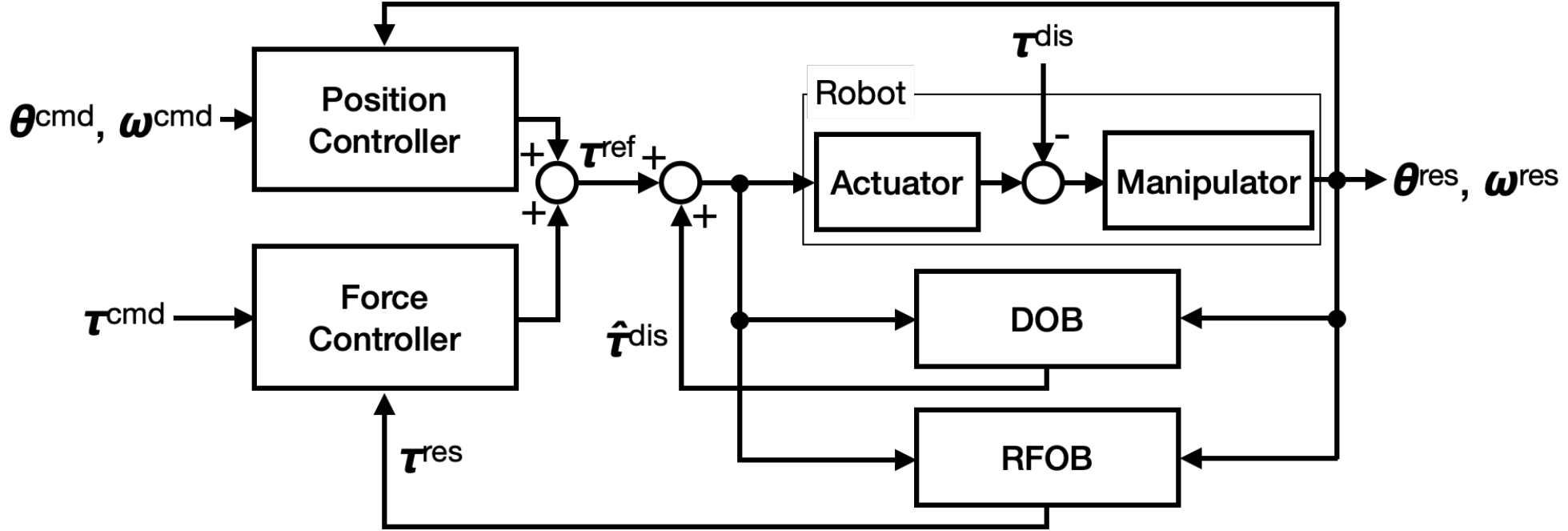}
          \caption{Hybrid controller with DOB and RFOB.}
          \label{fig:controller}
      \end{figure}

      Each joint of the manipulator was controlled using a hybrid position and force controller, as shown in figure~\ref{fig:controller}. 
      The superscript $dis$ depicts the disturbance, and the definitions of the other variables and superscripts are the same as in the aforementioned block diagrams.
      The controller consisted of a proportional--derivative (PD) controller for the joint angle and a proportional (P) controller for the joint force, equipped with a disturbance observer (DOB) and a reaction force observer (RFOB). 
      The parameters for the DOB/RFOB and gains for the controllers were those identified or set by Saigusa \textit{et al}.~\cite{CRANEX7params}.
      The sampling frequency of the robots was set to 500~Hz. 

  \subsection{Setup of Imitation Learning}
    Bilateral control-based imitation learning was used to evaluate the effects of real-world data augmentation.

    As mentioned earlier, datasets with simple duplication and changes in the speed of the command/response value of the human demonstration were also prepared to evaluate the effect of real-world responses through comparison with the proposed method.
    
\begin{figure}[tbhp]
      \centering
      \includegraphics[scale=0.25, bb = 0 0 955 285]{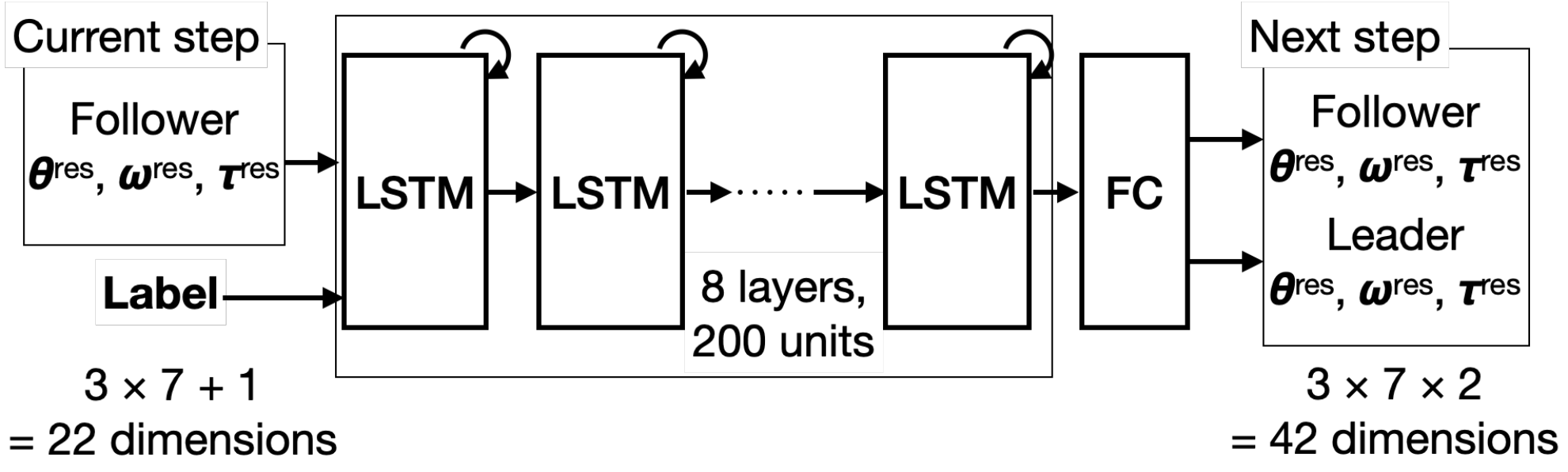}
      \caption{Configuration of NN model.}
      \label{fig:model2}
    \end{figure}

    The sampling frequency of both the bilateral control and teaching--playback was set to 500~Hz.
    NN models were trained using data obtained through either teaching--playback or simply duplicating and altering the speed. 
    Each NN model consisted of eight layers, 200 units of long short-term memory (LSTM)~\cite{LSTM1, LSTM2} cells, and one fully connected layer, as shown in figure~\ref{fig:model2}.
    The input had 22 dimensions in total, seven each for the follower's joint angles, angular velocities, and torques, and one for the label, depending on the task.
    The output had 42 dimensions: the follower's joint angles, angular velocities, and torques in addition to those of the leader's, making the model a follower-to-follower/leader (F2FL)~\cite{F2FL} structure to ensure that the prediction was based on a variety of environments.
    Each input and output dimension was normalized, with the average set to 0 and the standard deviation set to 1.
    The frequency of the NN model was 50~Hz. Each sample of the obtained data, which was at 500~Hz, was downsampled and rearranged~\cite{rahmatizadeh2018} as ten samples. 
    In the evaluation, the control frequency of the robot remained at 500~Hz, whereas the command value inferred by the NN model only changed at 50~Hz.
    In addition, for the training data, Gaussian noise with a size of 0.01 times the standard deviation was added to the input to prevent overfitting.
    The learning rate was $1.0~\times~10^{-4}$ and the models were trained for 5000 epochs.

  \subsection{Setup of Tasks}
    The experiments were conducted on two tasks, pick-and-place and wiping, using different data and environments for training and evaluation. 
    \subsubsection{Pick-and-Place Task}
\begin{figure}[tbhp]
        \centering
        \subfloat[Initial State]{%
          \resizebox*{3cm}{!}{\includegraphics[scale=0.1, bb = 0 0 555 633]{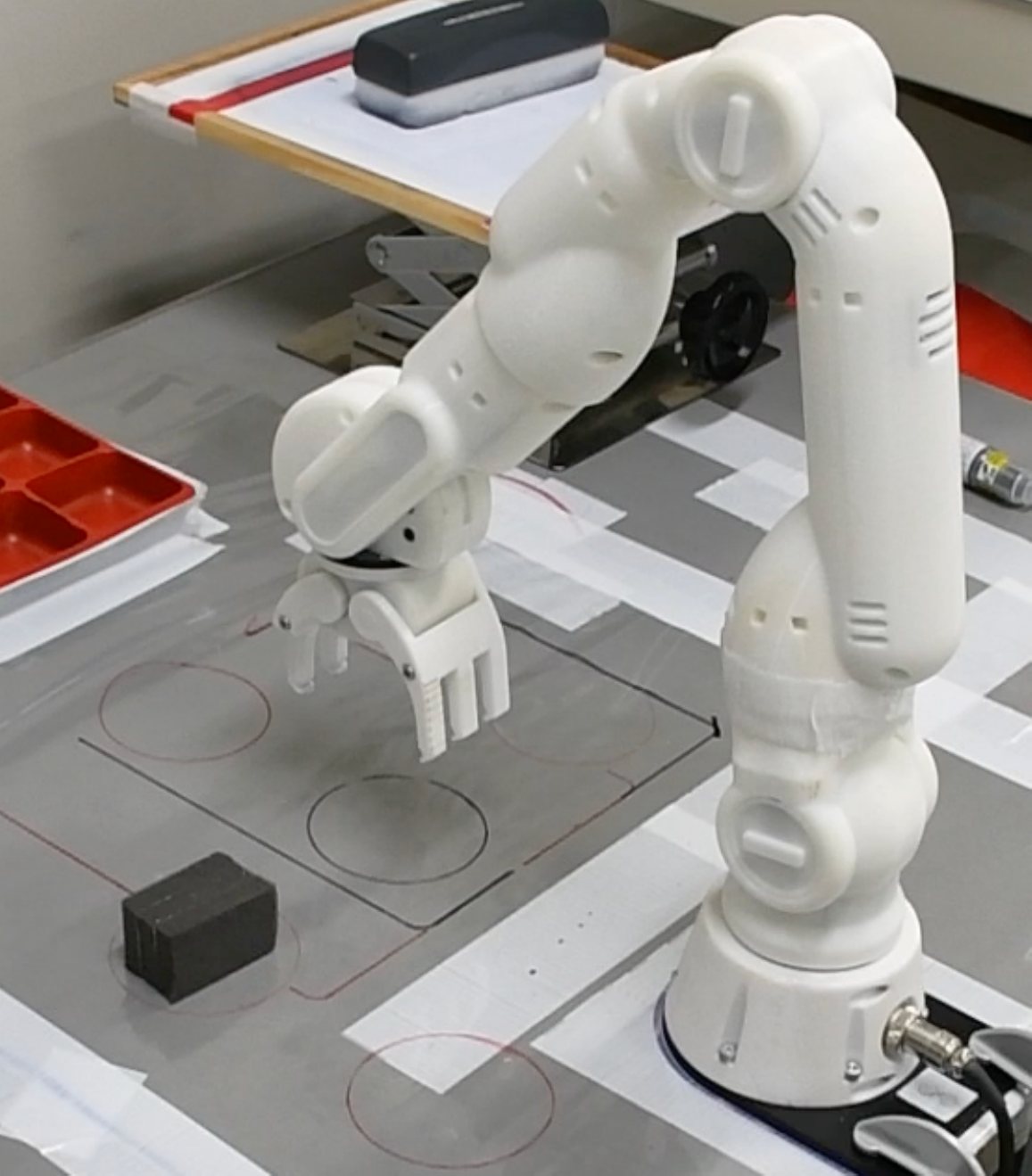}}
        }
        \hspace{5pt}
        \subfloat[Grasping]{%
          \resizebox*{3cm}{!}{\includegraphics[scale=0.1, bb = 0 0 581 612]{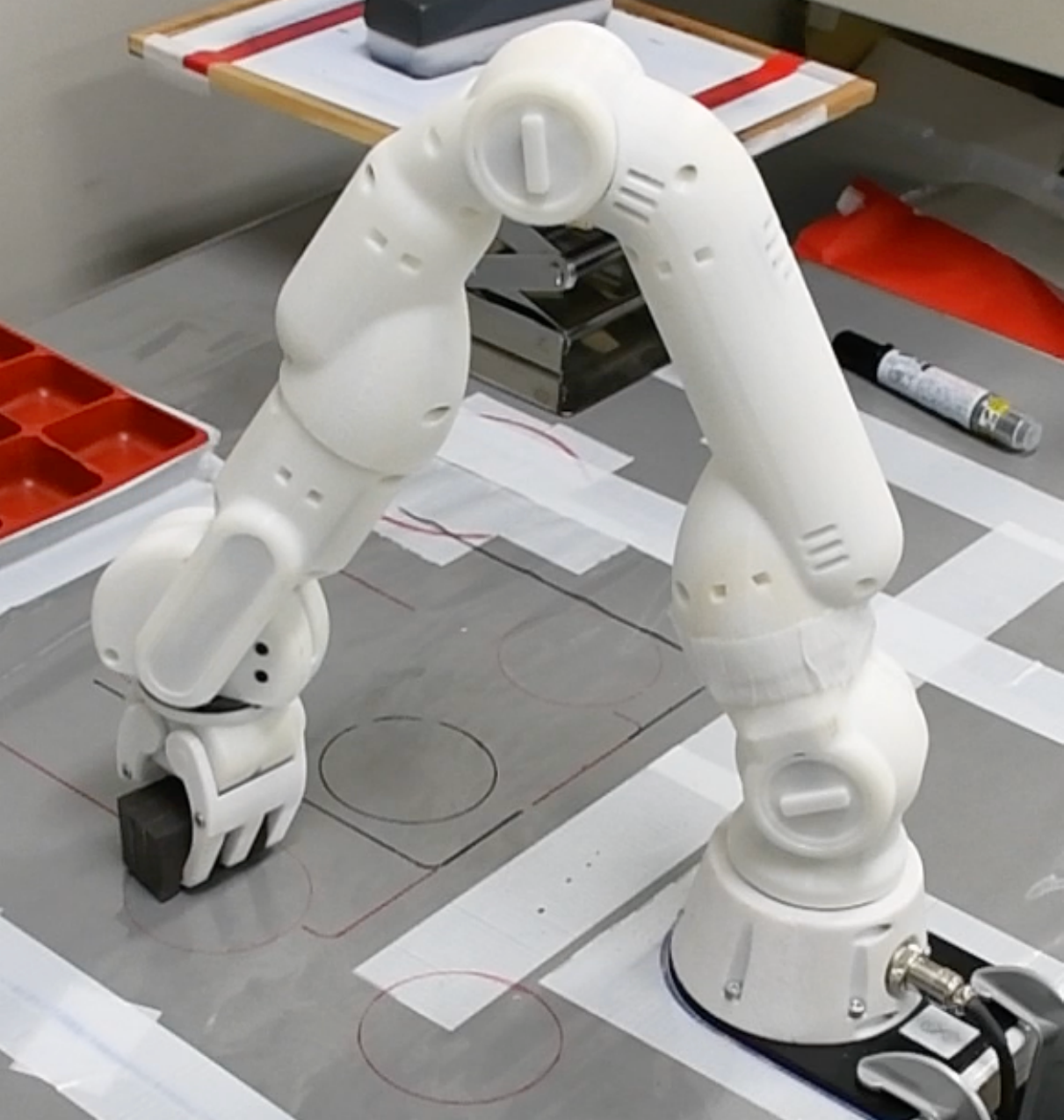}}
        }
        \hspace{5pt}
        \subfloat[Carrying]{%
          \resizebox*{3cm}{!}{\includegraphics[scale=0.1, bb = 0 0 532 632]{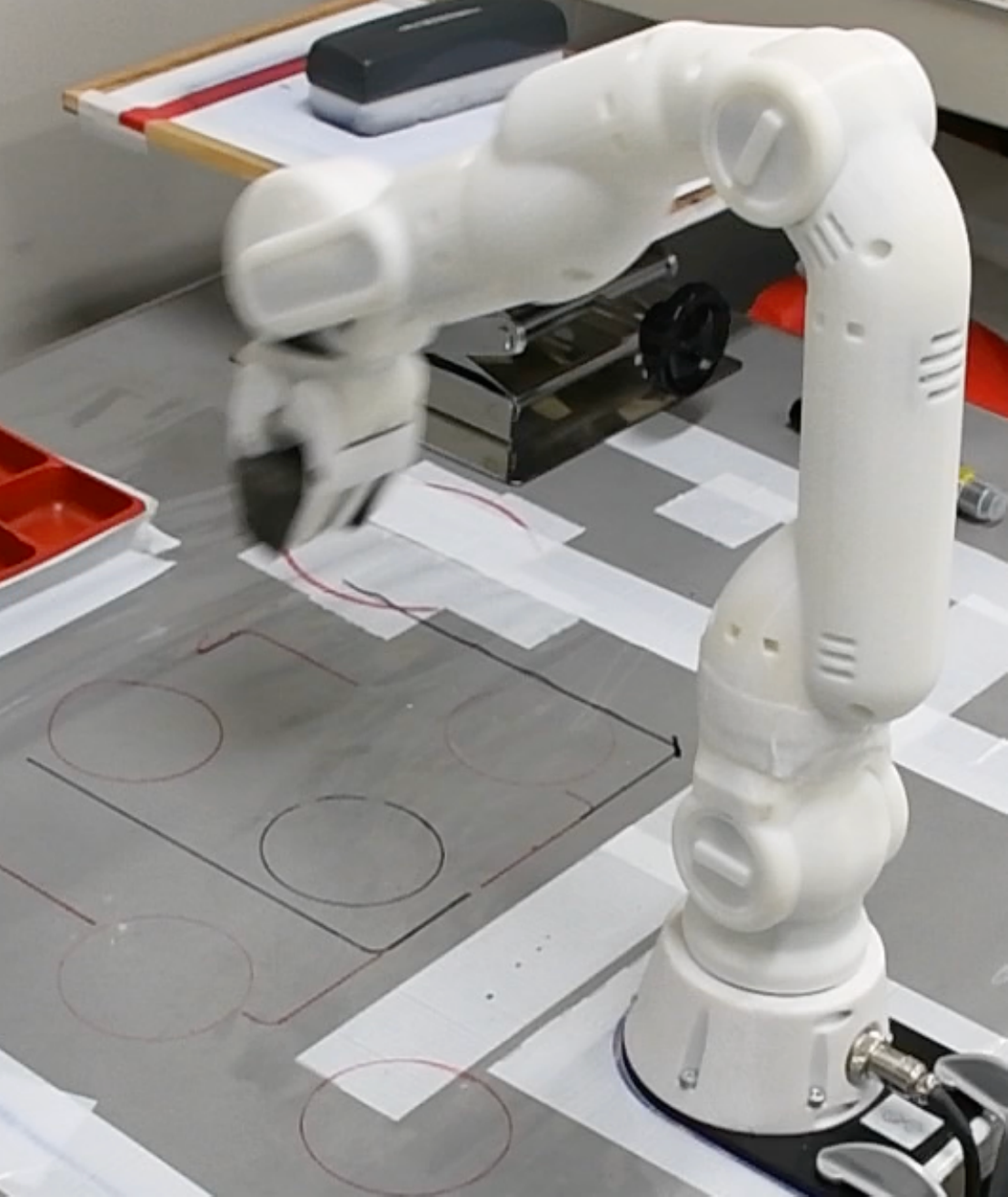}}
        }
        \hspace{5pt}
        \subfloat[Placing]{%
          \resizebox*{3cm}{!}{\includegraphics[scale=0.1, bb = 0 0 542 647]{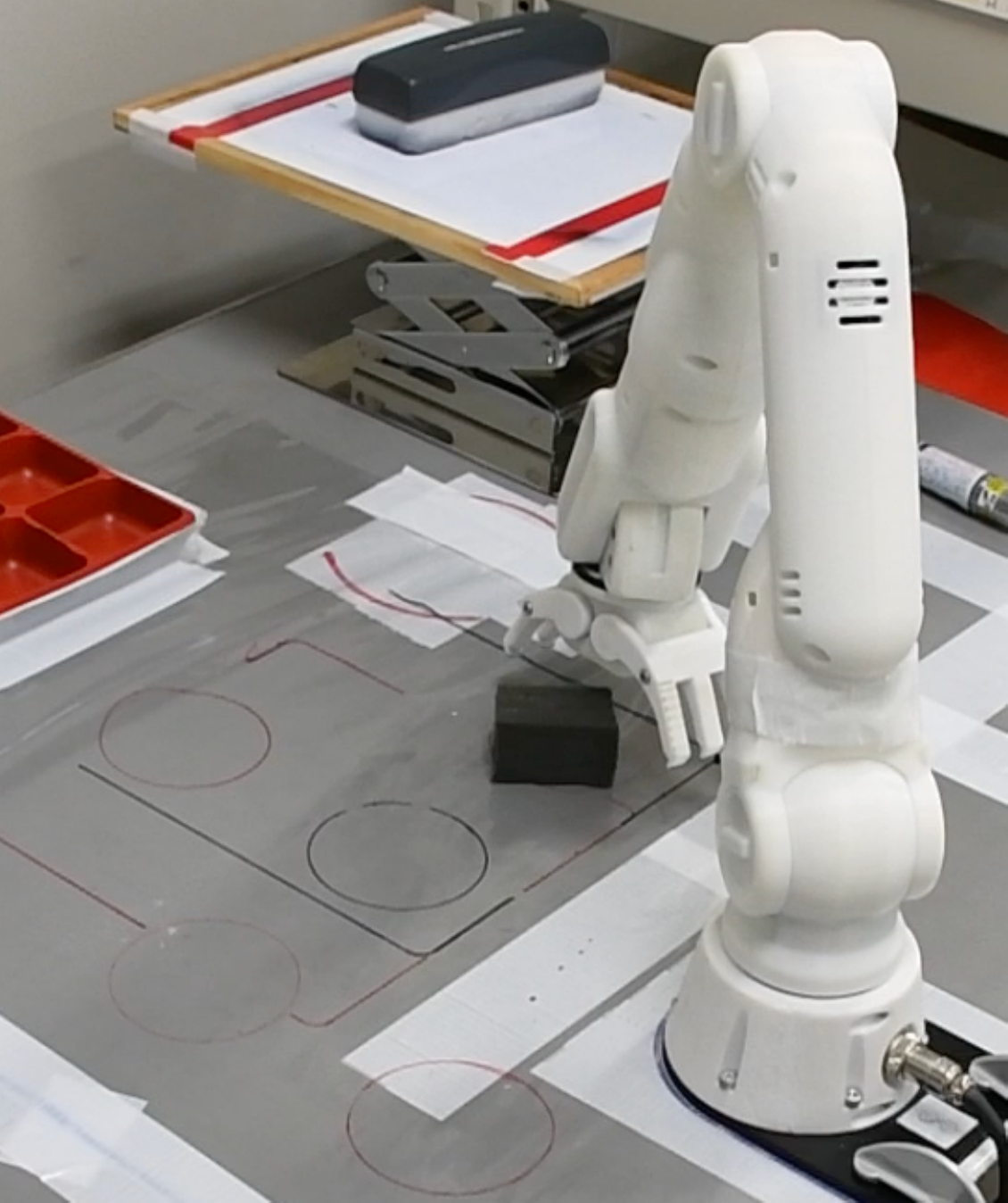}}
        }
        \caption{Procedure of the pick-and-place task.}
        \label{fig:pptask}
      \end{figure}

      In the pick-and-place task, as depicted in figure~\ref{fig:pptask}, the robot first grasped an object placed within the circle on the left side, carried it, and then placed it within the circle on the right side. 
      A trial was considered a failure when the entire object dropped out of the circle or the task was not completed within 40 s. 

\begin{figure}[tbhp]
        \centering
        \subfloat[Hard Sponge]{%
          \resizebox*{3cm}{!}{\includegraphics[scale=0.04, bb = 0 0 890 579]{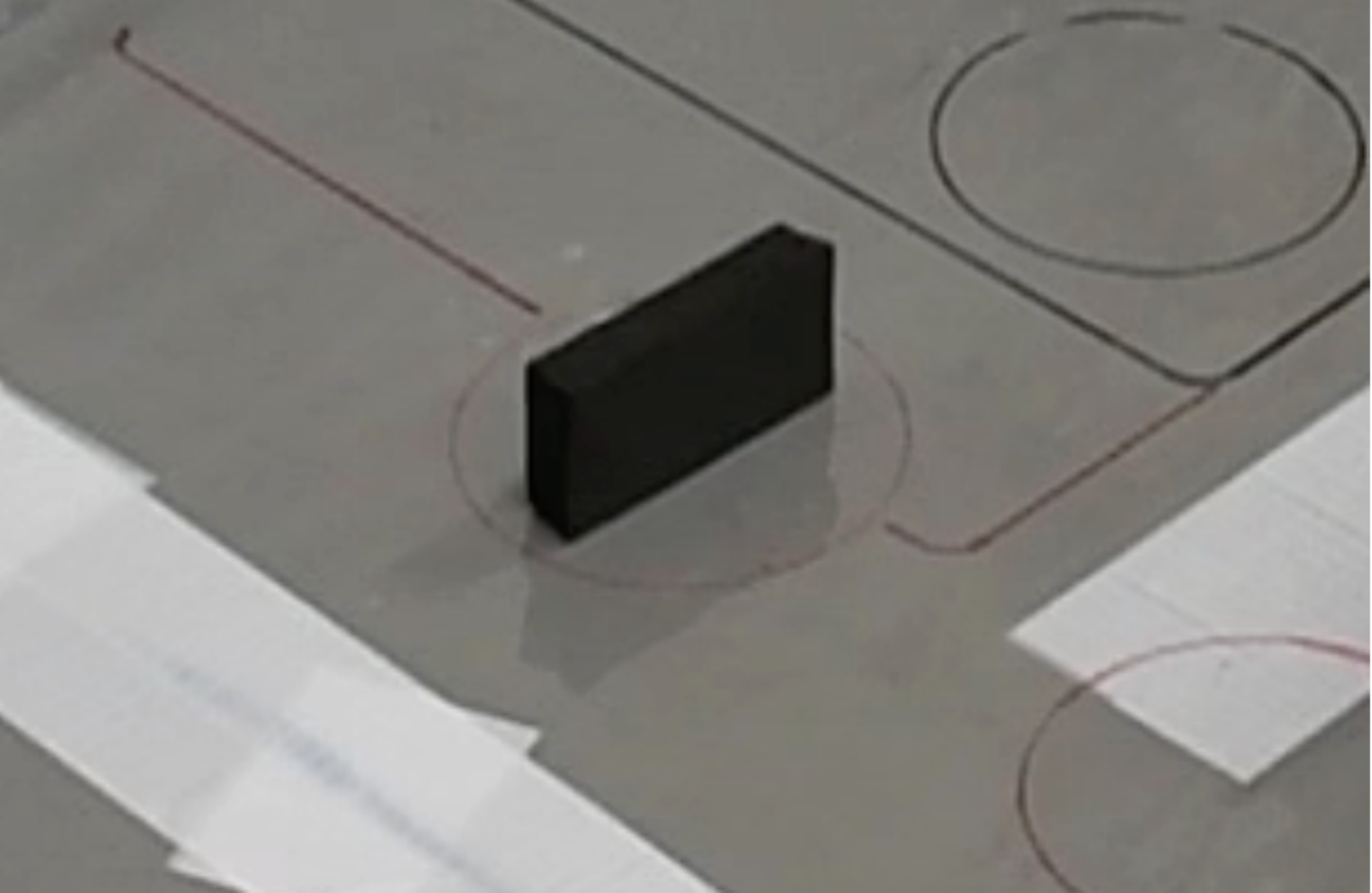}}
        }
        \hspace{5pt}
        \subfloat[Soft Sponge]{%
          \resizebox*{3cm}{!}{\includegraphics[scale=0.04, bb = 0 0 816 579]{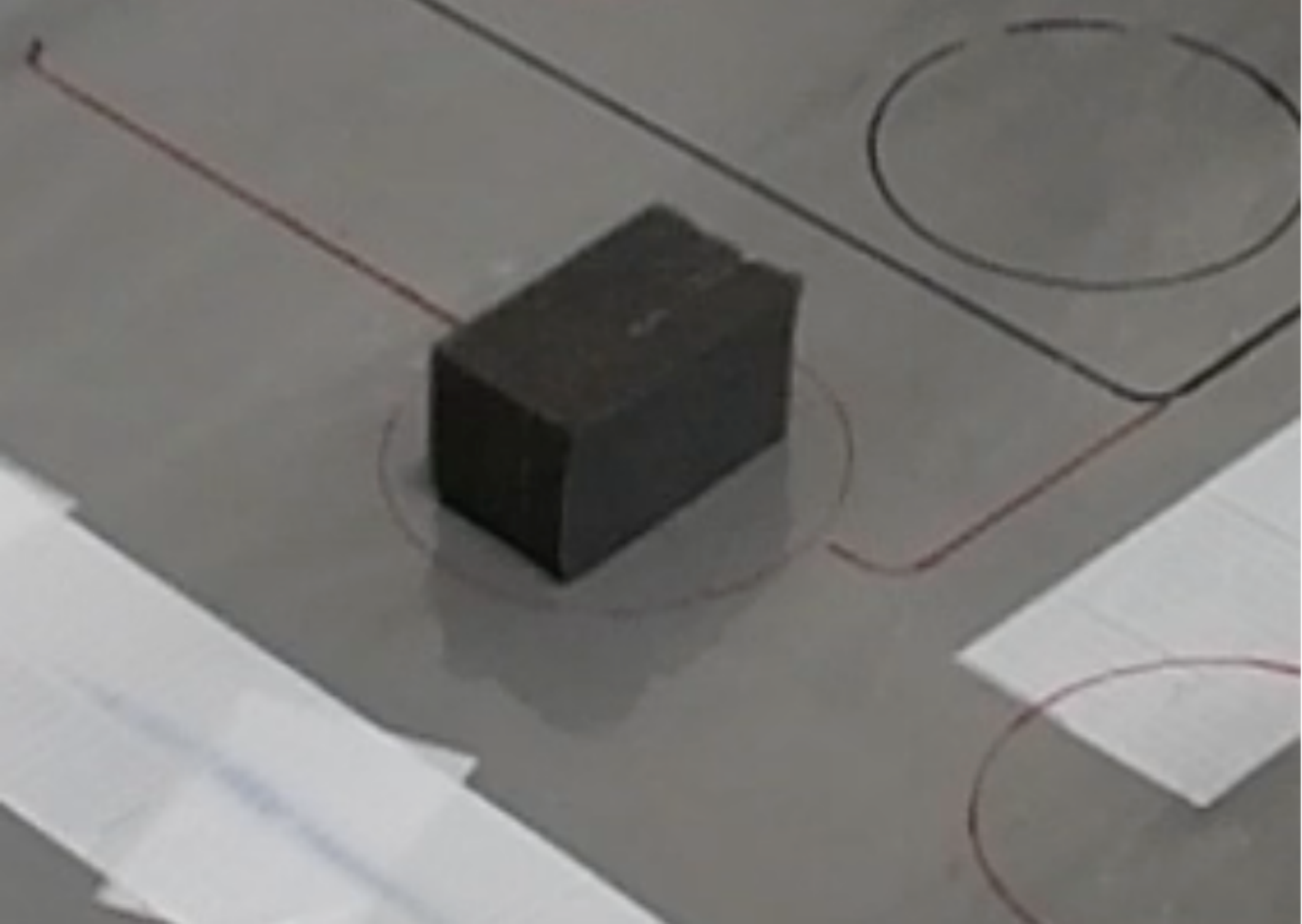}}
        }
        \hspace{5pt}
        \subfloat[Electric Tap]{%
          \resizebox*{3cm}{!}{\includegraphics[scale=0.04, bb = 0 0 833 579]{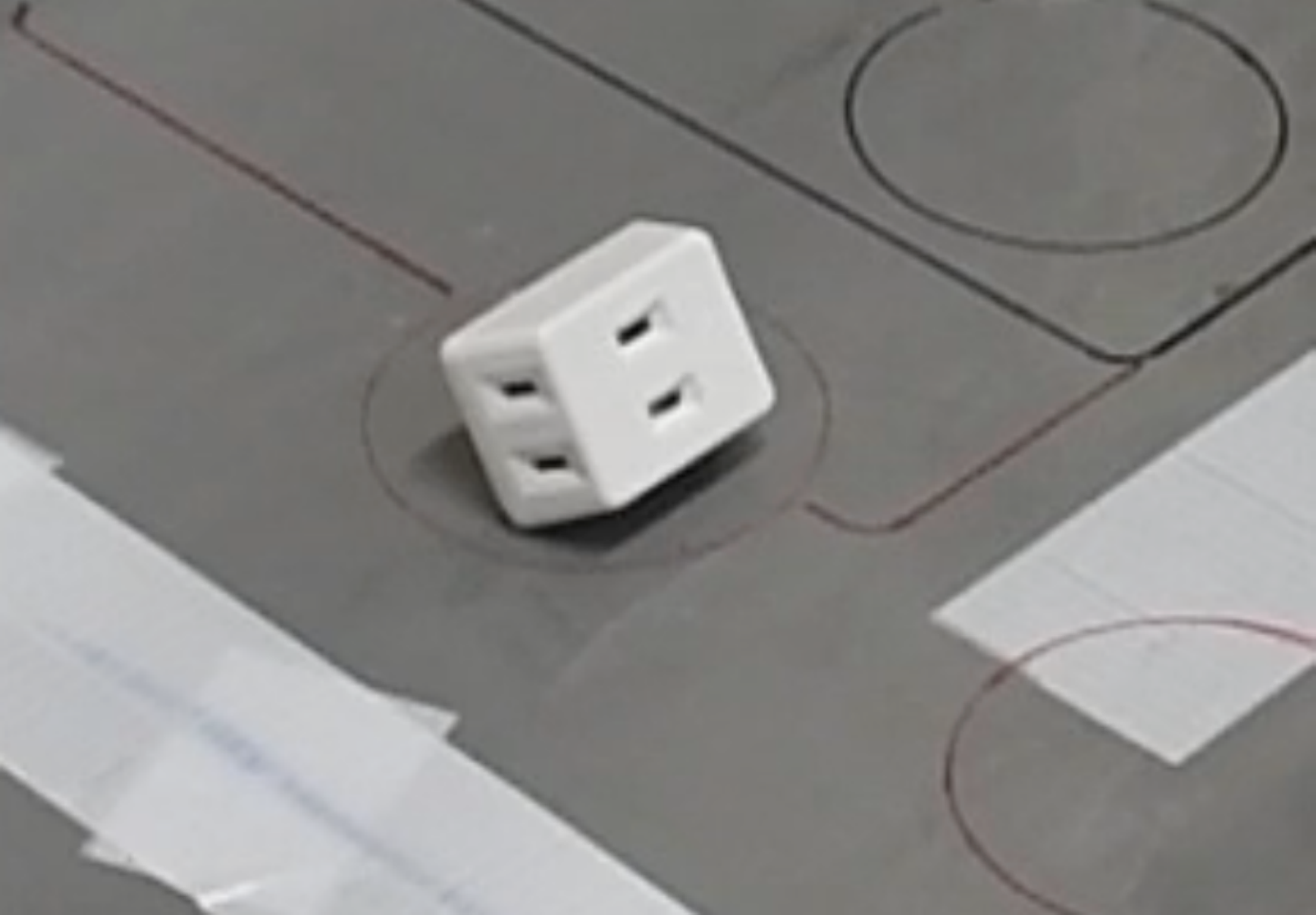}}
        }
        \hspace{5pt}
        \subfloat[Cloth]{%
          \resizebox*{3cm}{!}{\includegraphics[scale=0.04, bb = 0 0 864 579]{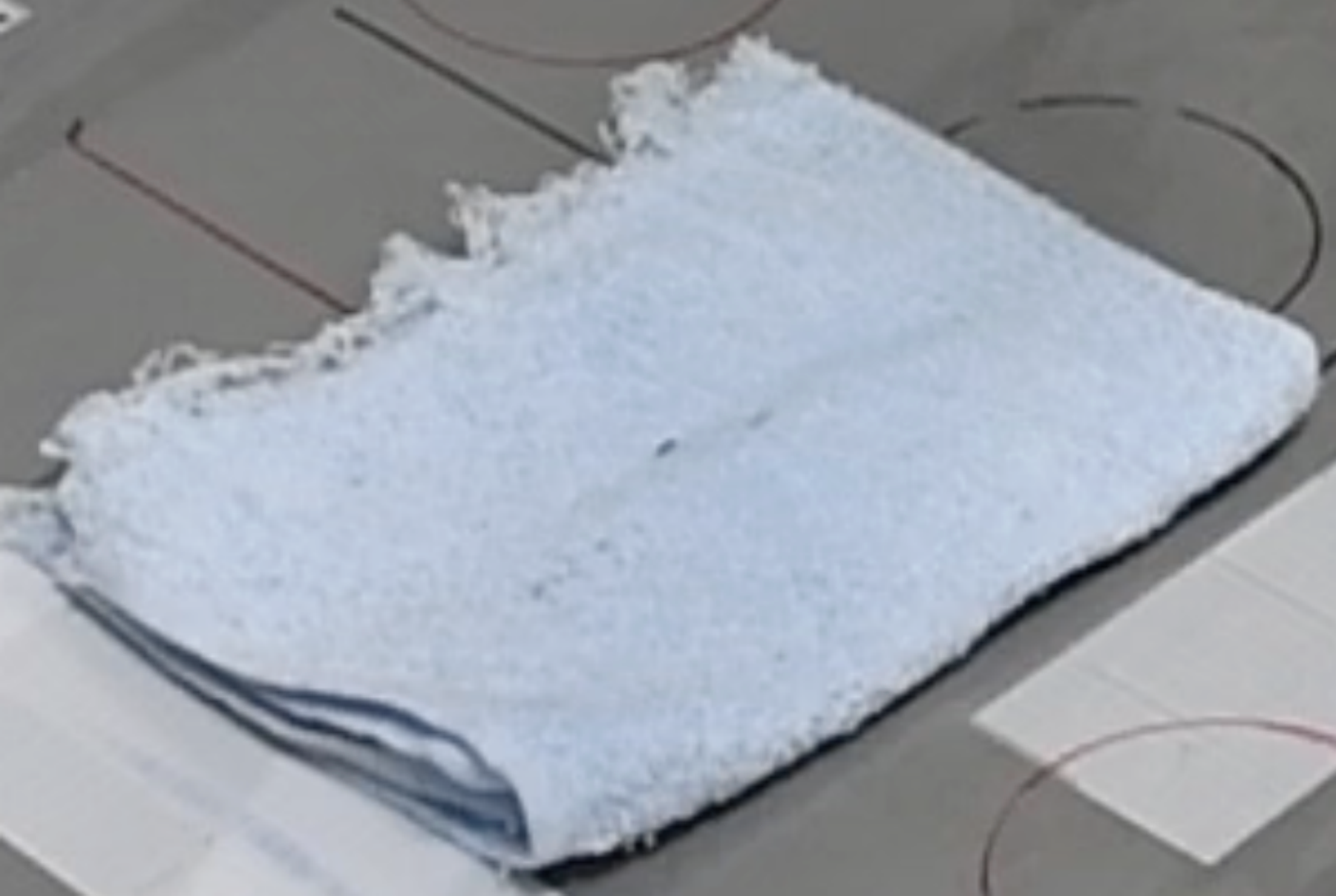}}
        }
        \caption{Objects for the pick-and-place task.}
        \label{fig:ppobj}
      \end{figure}

      Four objects, as shown in figure~\ref{fig:ppobj}, were prepared for this task. 
      Two sponges, a hard and soft one, were utilized for both training and evaluation, whereas the other two objects were selected only for evaluation.
      The hard sponge was a natural rubber sponge approximately 1~cm thick, whereas the soft sponge was a urethane sponge approximately 3~cm thick; both were selected from the objects of Yamane \textit{et al},~\cite{yamaneral}. 
      An electric tap and cloth were selected from everyday items to simulate a much wider range of impedance and shape uncertainties.

      In the initial data collection phase, two demonstrations were conducted using four-channel bilateral control. 
      One was with the hard sponge, which took 6.6~s, and the other with the soft sponge, which took 7.0~s. 
      For data augmentation, the environmental responses from playbacks were collected at 0.5, 1, and 2 times the speed of these two sponges, 10 times each, totaling 60 playbacks.
      The data obtained from the playback were labeled by the length of time when the angle of the end effector was smaller than 3.7~rad, which means that the hand was closed.
      Seven samples of each sponge and speed were used for training, and the other three samples were used for validation.
      In the trial, labels of 1.5, 3, 6, 9, 12, and 15~s were used, and unlearned objects, electric taps, and cloth, as depicted in figure~\ref{fig:ppobj}, were added. 

    \subsubsection{Wiping Task}

\begin{figure}[tbhp]
        \centering
        \subfloat[Initial State]{%
          \resizebox*{3cm}{!}{\includegraphics[scale=0.36, bb = 0 0 712 905]{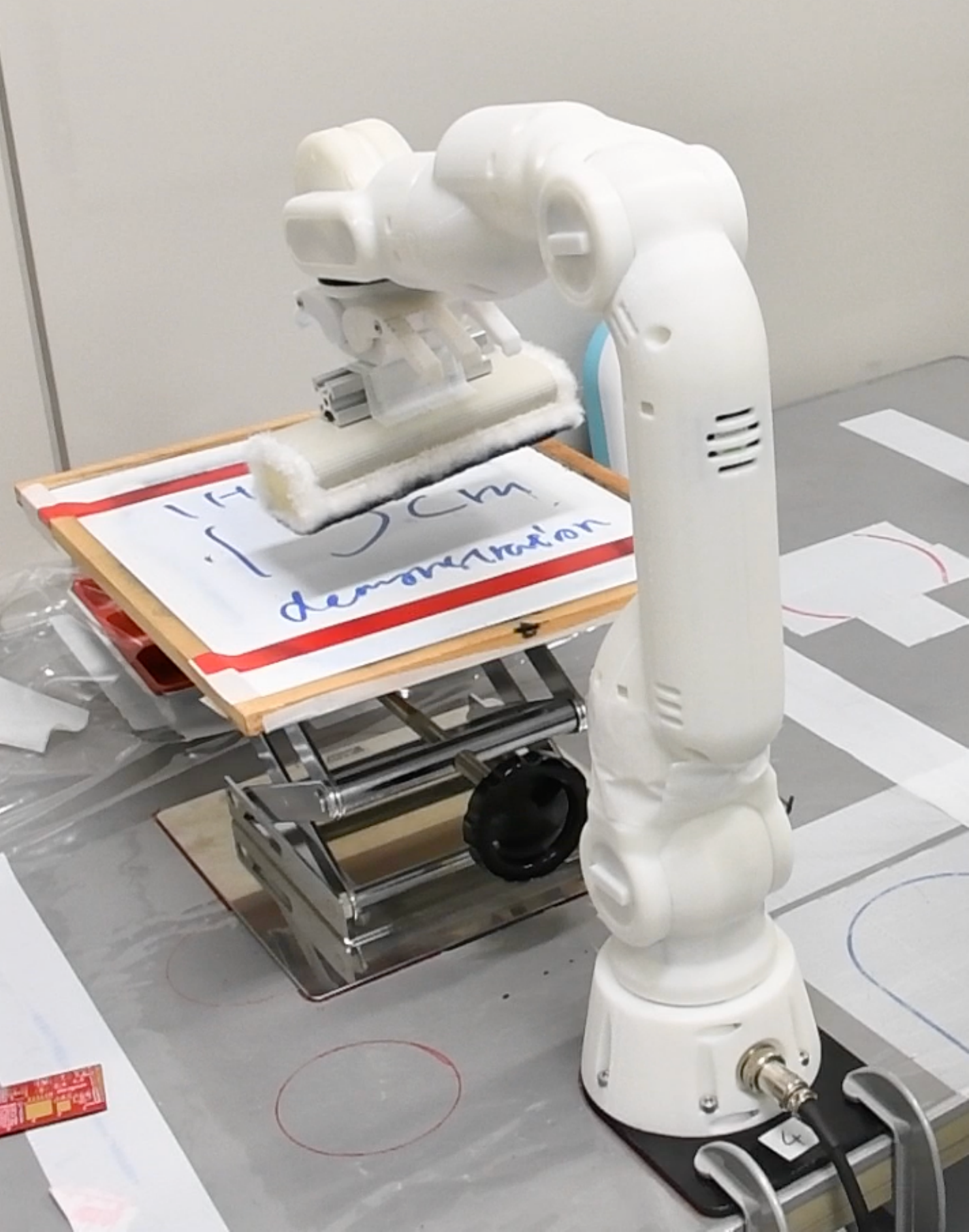}}
        }
        \hspace{5pt}
        \subfloat[Pressing]{%
          \resizebox*{3cm}{!}{\includegraphics[scale=0.36, bb = 0 0 712 906]{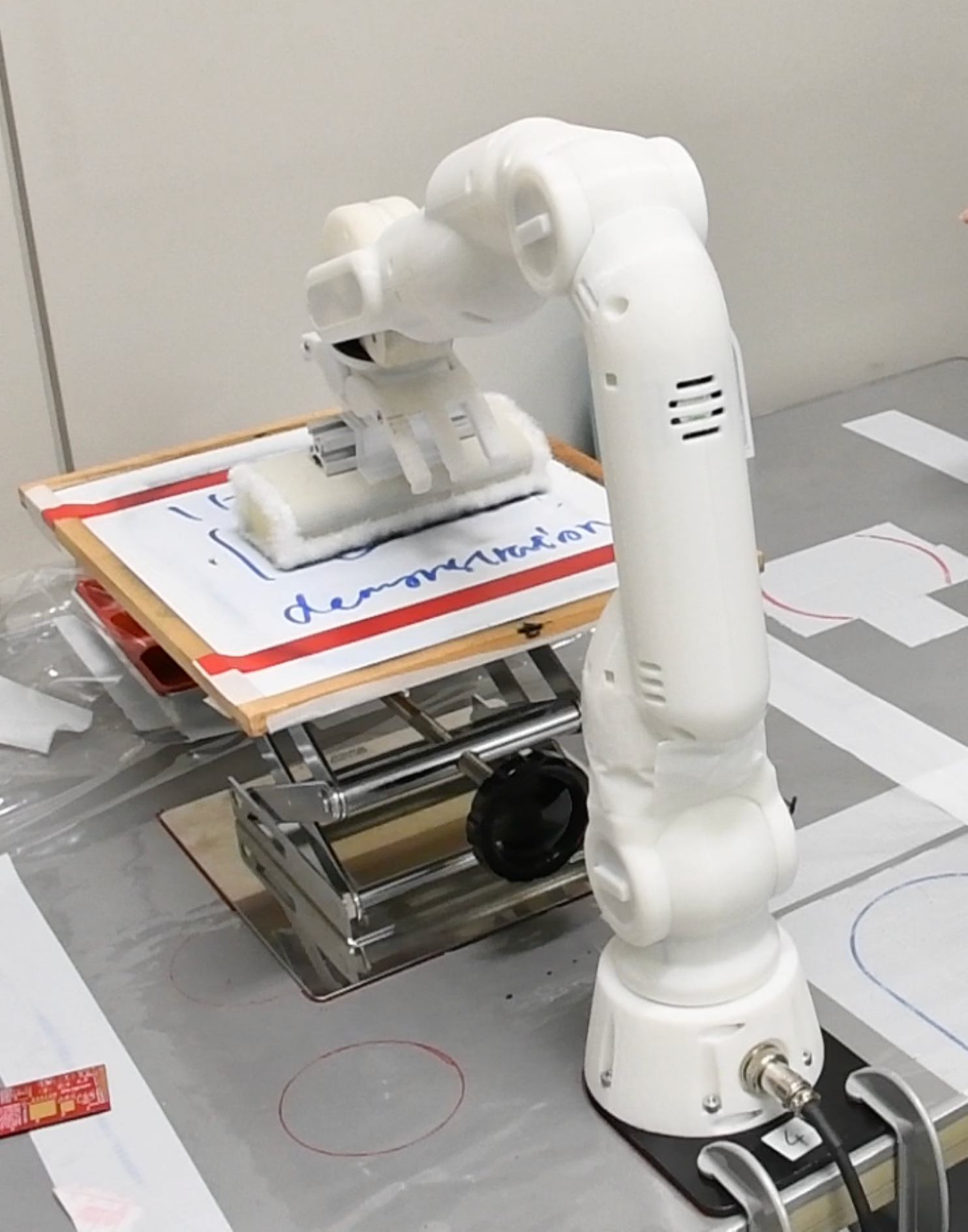}}
        }
        \hspace{5pt}
        \subfloat[Grasping]{%
          \resizebox*{3cm}{!}{\includegraphics[scale=0.36, bb = 0 0 711 906]{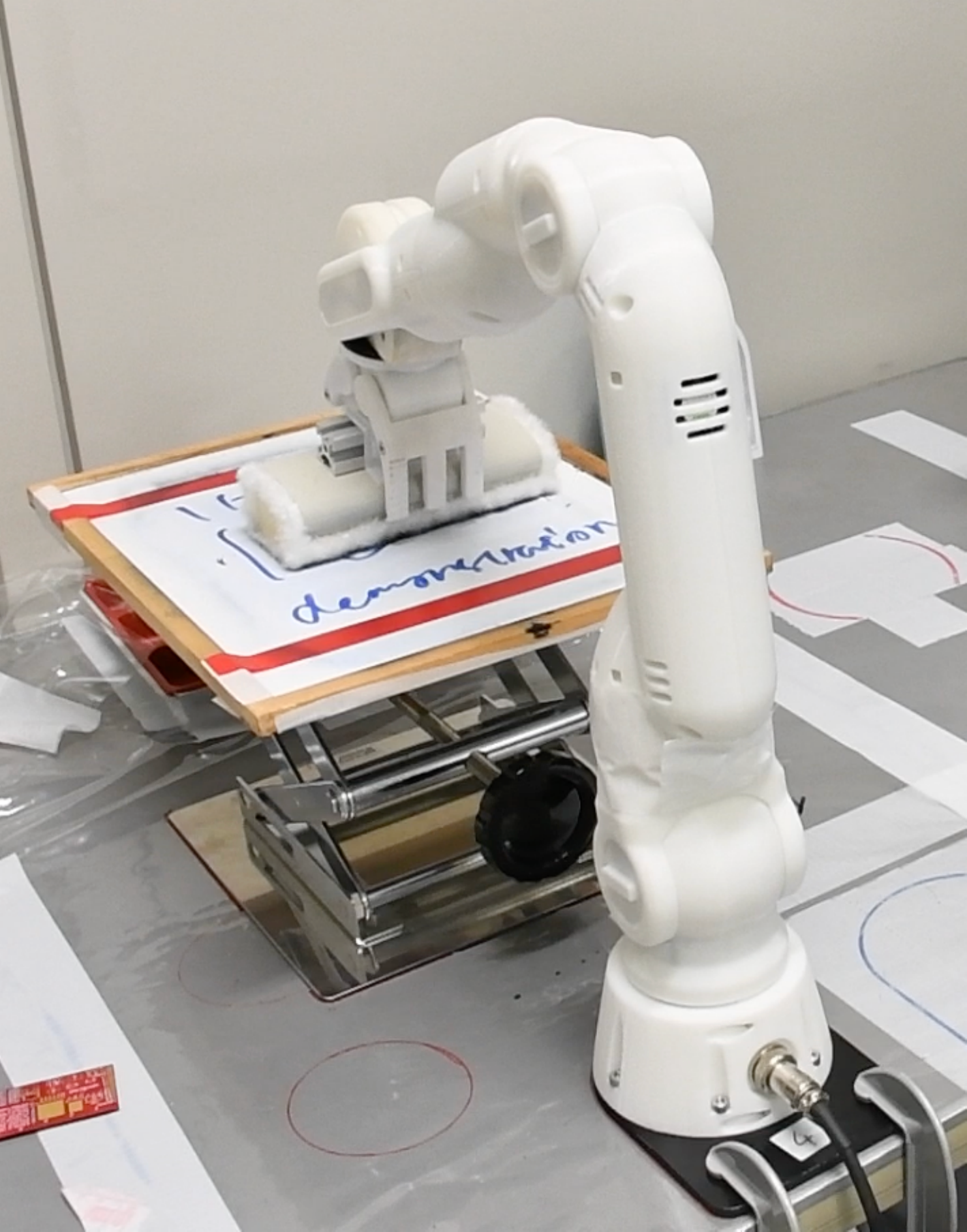}}
        }
        \hspace{5pt}
        \subfloat[Wiping]{%
          \resizebox*{3cm}{!}{\includegraphics[scale=0.36, bb = 0 0 710 904]{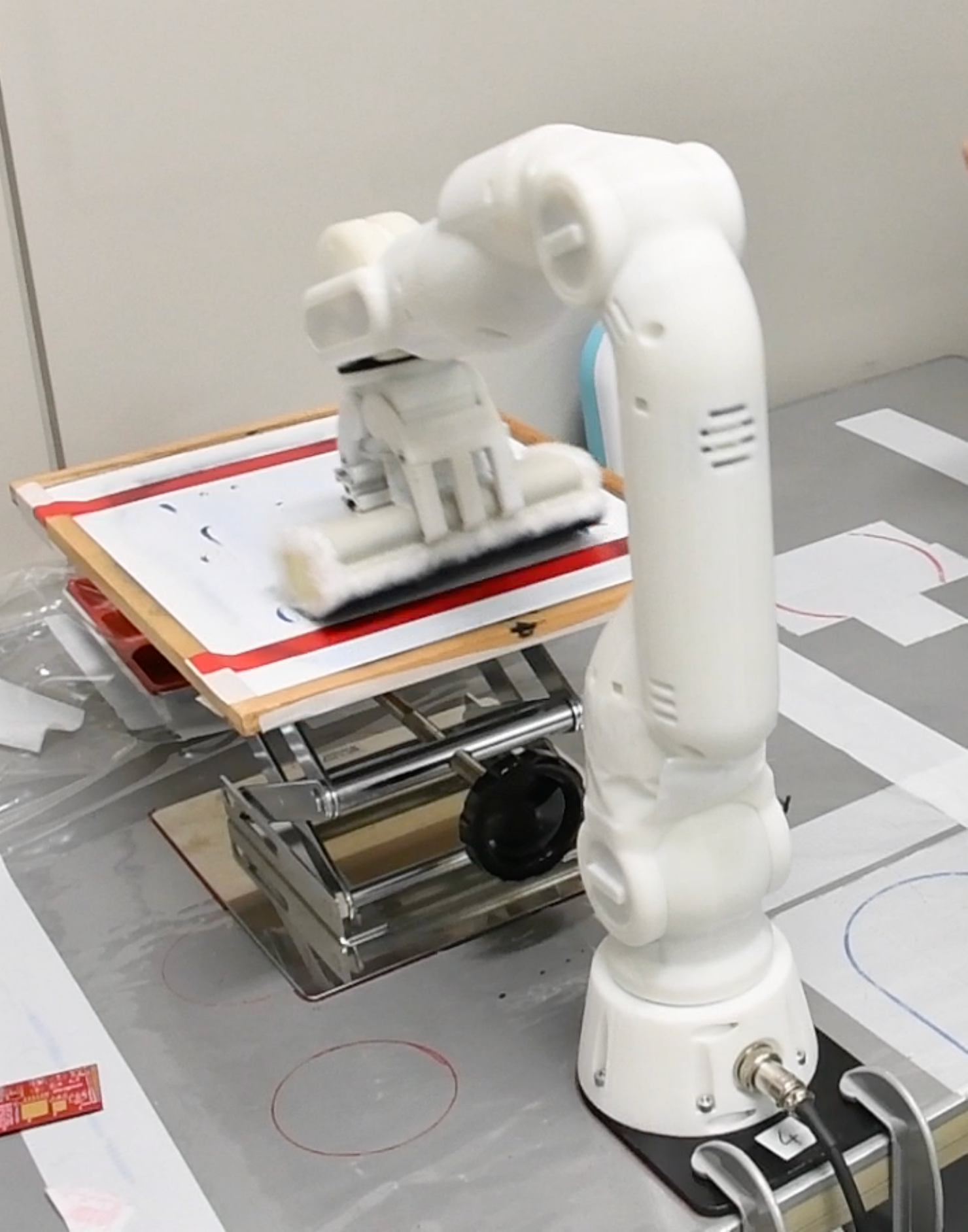}}
        }
        \caption{Procedure of the wiping task.}
        \label{fig:wipetask}
      \end{figure}

      In the wiping task, depicted in figure~\ref{fig:wipetask}, the robot first pressed an eraser towards a whiteboard placed horizontally on a desk, grasped the eraser, and then wiped the board continuously at different frequencies. 
      The height of the surface was set at either 12 or 15~cm from the desk. 
      The eraser was equipped with an aluminum frame for adjustment of height and was loosely attached to the robot's hand.
      Trials were considered to have failed when the eraser lost contact with the whiteboard or when the robot did not proceed to wiping. 

      In the data collection phase, two demonstrations were conducted using four-channel bilateral control. One was at a height of 12~cm, and the other was at 15~cm.
      All demonstrations were performed at a frequency of 1.0~Hz while listening to the sound of a metronome set at 120 beats per minute.

      In data augmentation, both demonstrations of the two heights were played back with 0.5, 1 and 1.5 times the speed, five times each.
      A total of 30 playbacks were collected, each with environmental reaction from pressing, grasping, and 40 s of wiping.
      For each height and speed, three playbacks were utilized as training data, and two were used for validation.
      The gathered data were labeled by the frequency of motion multiplied by the speed of playbacks, that is, 0.5, 1, or 1.5~Hz.

      In the trial, labels of 0.25, 0.5, 0.75, 1, 1.25, 1.5, and 1.75~Hz were used.
      Each trial lasted 50 s and included pressing, grasping, and wiping. 

  \subsection{Result}
    \subsubsection{Pick-and-Place Task}
      Tables~\ref{ppmocopy}~and~\ref{ppaug3} depict the success rates of the trials in the pick-and-place task. 
      Labels between 4.5 and 12~s were interpolated for the NN models, and the others were extrapolated.
      Real-world data augmentation outperformed the success rate, except for five conditions: hard sponge and electric tap labeled 1.5~s, electric tap labeled 9~s, and hard sponge and cloth labeled 15~s. 
      The overall success rate differed more than 20\% and more than 50\% for interpolation, given that the change in speed equaled to a range of 3.3--14~s. 

\begin{figure}[tbhp]
        \centering
        \subfloat[With real-world data augmentation]{%
          \resizebox*{6cm}{!}{\includegraphics[scale=0.6, bb = 0 0 363 215]{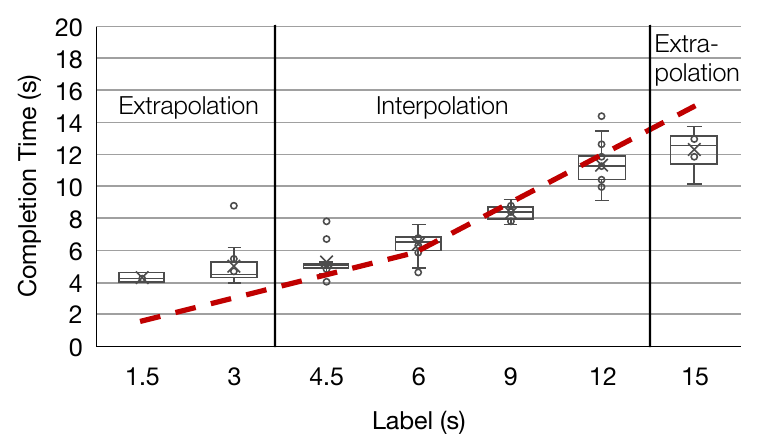}}
        }
        \hspace{5pt}
        \subfloat[Without real-world data augmentation]{%
          \resizebox*{6cm}{!}{\includegraphics[scale=0.6, bb = 0 0 363 215]{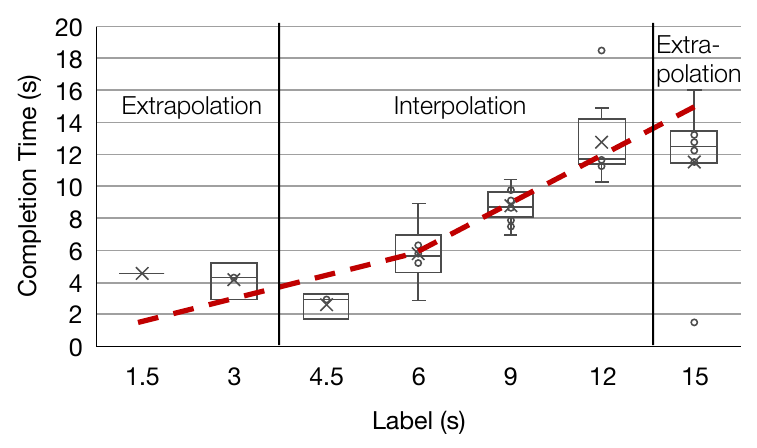}}
        }
        \caption{Actual completion time in successful trials of pick-and-place task. Matching the label (red dashed line) was better.}
        \label{fig:ppshoyo}
      \end{figure}

      The actual times consumed for each successful trial are shown in figure~\ref{fig:ppshoyo}. 
      According to the figure, real-world data augmentation resulted in smaller leaps from the given labels at slower conditions, such as 12 and 15~s, and a stronger correlation to the given labels, even for extrapolation. 
      This shows that the application of real-world data augmentation contributed to higher accuracy and stability, particularly in slower conditions, and the limitations of extrapolation through the NN model to faster motion, particularly twice as fast as taught.

      \begin{table}
        \tbl{Task success rate of the pick-and-place task, trained with real-world playbacks at 0.5x, 1x, and 2x the speed of human demonstrations.}
        {
          \begin{tabular}{lccccccc|cc} 
            \toprule & \multicolumn{2}{l}{Label} \\ \cmidrule{2-8}
            Object       & 1.5 s            & 3 s              & 4.5 s            & 6 s              & 9 s              & 12 s             & 15 s            &4.5--12 s& Overall           \\ \midrule
            Hard Sponge  & \emph{1/5(20\%)} & \emph{4/5(80\%)} & \emph{3/5(60\%)} & \emph{3/5(60\%)} & \emph{5/5(100\%)}& \emph{5/5(100\%)}& 0/5(0\%)        &\emph{16/20(80\%)}       &\emph{21/35(60\%)} \\
            Soft Sponge  & \emph{1/5(20\%)} & \emph{3/5(60\%)} & \emph{5/5(100\%)}& \emph{4/5(80\%)} & \emph{5/5(100\%)}& \emph{4/5(80\%)} & \emph{3/5(60\%)}&\emph{18/20(90\%)}&\emph{25/35(71\%)} \\
            Electric Tap & \emph{0/5(0\%)}  & \emph{2/5(40\%)} & \emph{4/5(80\%)} & \emph{5/5(100\%)}& 4/5(80\%)        & \emph{5/5(100\%)}& \emph{3/5(60\%)}&\emph{18/20(90\%)}&\emph{23/35(66\%)} \\
            Cloth        & \emph{1/5(20\%)} & \emph{3/5(60\%)} & \emph{3/5(60\%)} & \emph{5/5(100\%)}& \emph{5/5(100\%)}& \emph{5/5(100\%)}& 0/5(0\%)        &\emph{18/20(90\%)}&\emph{22/35(63\%)} \\ \midrule
            Overall      & \emph{3/20(15\%)}&\emph{12/20(60\%)}&\emph{15/20(75\%)}&\emph{19/20(95\%)}&\emph{19/20(95\%)}&\emph{19/20(95\%)}& 6/20(30\%)      &\emph{70/80(88\%)}&\emph{74/140(53\%)} \\ \bottomrule
        \end{tabular}
        }
        \label{ppmocopy}
      \end{table}
      \begin{table}
        \tbl{Task success rate of the pick-and-place task, trained with human demonstrations simply duplicated and changed in speed.}
        {
          \begin{tabular}{lccccccc|cc} 
            \toprule & \multicolumn{2}{l}{Label} \\ \cmidrule{2-8}
            Object       & 1.5 s            & 3 s       & 4.5 s     & 6 s       & 9 s              & 12 s      & 15 s             &4.5--12 s& Overall\\ \midrule
            Hard Sponge  & \emph{1/5(20\%)} & 1/5(20\%) & 1/5(20\%) & 2/5(40\%) & 3/5(60\%)        & 4/5(80\%) & \emph{3/5(60\%)} & 10/20(50\%)             & 15/35(43\%)\\
            Soft Sponge  & 0/5(0\%)         & 0/5(0\%)  & 1/5(20\%) & 3/5(60\%) & 3/5(60\%)        & 1/5(20\%) & 2/5(40\%)        &  8/20(40\%)             & 10/35(29\%)\\
            Electric Tap & \emph{0/5(0\%)}  & 0/5(0\%)  & 0/5(0\%)  & 0/5(0\%)  & \emph{5/5(100\%)}& 2/5(40\%) & 1/5(20\%)        &  7/20(35\%)             & 8/35(23\%)\\
            Cloth        & 0/5(0\%)         & 2/5(40\%) & 1/5(20\%) & 1/5(20\%) & 2/5(40\%)        & 1/5(20\%) & \emph{2/5(40\%)} &  5/20(25\%)             & 9/35(26\%)\\ \midrule
            Overall      & 1/20(5\%)        & 3/20(15\%)& 3/20(15\%)&6/20(30\%) & 13/20(65\%)      &8/20(40\%) & \emph{8/20(40\%)}& 25/80(31\%)             & 42/140(30\%)\\ \bottomrule
        \end{tabular}
        }
        \label{ppaug3}
      \end{table}
      
    \subsubsection{Wiping Task}
      The results of the wiping task are presented in tables~\ref{wpmocopy}~to~\ref{wpaug3}. 
      Labels between 0.5 and 1.5~Hz were interpolation for the NN models, and others were extrapolated.
      The overall success rate with real-world data augmentation was slightly higher than that without, whereas at low frequencies, real-world data augmentation had a higher success rate at 15~cm and a lower success rate at 12~cm. 
      The main reason for failure without real-world data augmentation was the initiation of periodical movement before pressing, resulting in the eraser bumping towards the whiteboard.
      As no failures of this type were observed with real-world data augmentation, this indicated that the proposed method enabled learning the procedure of the task according to force feedback. 
      However, failures with real-world data augmentation are primarily sudden losses of contact force or repeated pressing and grasping. 
      Because these failures were more frequent at low heights, this indicated insufficient height detection.    

\begin{figure}[tbhp]
        \centering
        \subfloat[With real-world data augmentation]{%
          \resizebox*{6cm}{!}{\includegraphics[scale=0.6, bb = 0 0 363 215]{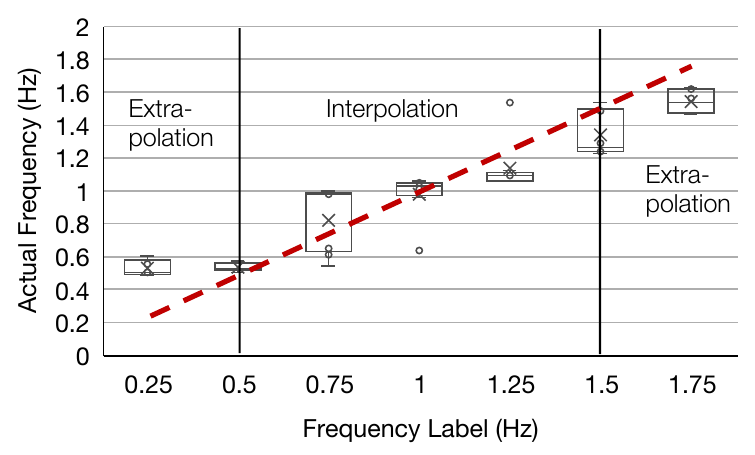}}
        }
        \hspace{5pt}
        \subfloat[Without real-world data augmentation]{%
          \resizebox*{6cm}{!}{\includegraphics[scale=0.6, bb = 0 0 363 215]{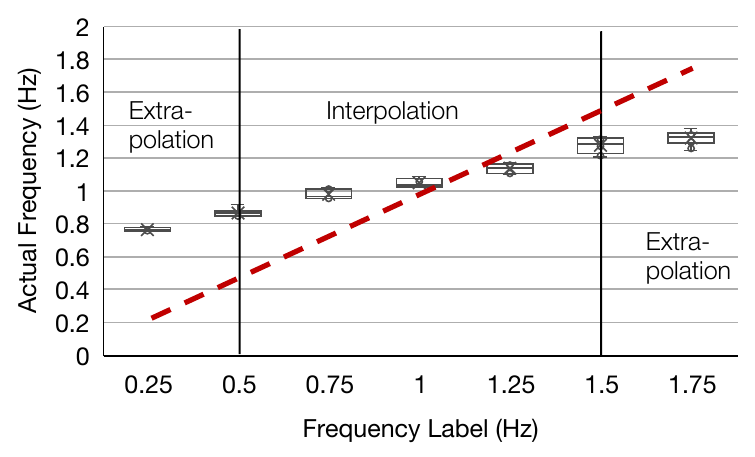}}
        }
        \caption{Actual frequency of wiping motion in successful trials of the wiping task, averaged for each trial. Matching the label (red dashed line) was better.}
        \label{fig:wpshoyo}
      \end{figure}

      Figure~\ref{fig:wpshoyo} depicts the actual frequency of wiping in successful trials, averaged for each trial.
      Given the fluctuation in frequency at the beginning of motion in human demonstrations and the lack of a metronome-like input to the NN model, small differences can be tolerated.
      From the data shown in the figure, real-world data augmentation contributed to the accuracy along the given label at both high and low frequencies, in contrast to the model without the proposed method, which remained between 0.7 and 1.4~Hz.
      A limited frequency range, larger distributions and outliers in the proposed method indicate the necessity for fine-tuning and a wider range of training data, which are available by more playbacks in diverse settings.
  
      \begin{table}
        \tbl{Task success rate of the wiping task, trained with real-world playbacks of 0.5x, 1x, and 1.5x the speed of human demonstrations at 1 Hz.}
        {\resizebox*{20cm}{!}{
          \begin{tabular}{lccccccccc|cc} 
            \toprule & \multicolumn{2}{l}{Label} \\ \cmidrule{2-10}
            Height & 0.25 Hz           & 0.5 Hz            & 0.75 Hz            & 1 Hz               & 1.25 Hz            & 1.5 Hz          & 1.75 Hz           &0.5--1.5 Hz       & Overall\\ \midrule
            15 cm  & \emph{ 4/5(80\%)} & \emph{ 4/5(80\%)} & \emph{ 5/5(100\%)} & \emph{ 5/5(100\%)} & \emph{ 5/5(100\%)}& \emph{5/5(100\%)}& \emph{5/5(100\%)} & \emph{24/25(96\%)}& \emph{33/35(94\%)}\\
            12 cm  &         1/5(20\%) &        1/5(20\%)  & \emph{ 4/5(80\%)}  &        3/5(60\%)   & \emph{ 4/5(80\%)} & \emph{ 4/5(80\%)}& \emph{5/5(100\%)} &       16/25(64\%) &       22/35(63\%)\\ \midrule
            Overall& \emph{5/10(50\%)} &        5/10(50\%) & \emph{ 9/10(90\%)} & \emph{8/10(80\%)}  & \emph{ 9/10(90\%)}& \emph{9/10(90\%)}&\emph{10/10(100\%)}& \emph{40/50(80\%)}& \emph{55/70(79\%)}\\ \bottomrule
        \end{tabular}}
        }
        \label{wpmocopy}
      \end{table}
      \begin{table}
        \tbl{Task success rate of wiping task, trained with human demonstrations simply duplicated and changed in speed.}
        {\resizebox*{20cm}{!}{
          \begin{tabular}{lccccccccc|cc} 
            \toprule & \multicolumn{2}{l}{Label} \\ \cmidrule{2-10}
            Height & 0.25 Hz          & 0.5 Hz          & 0.75 Hz          & 1 Hz             & 1.25 Hz          & 1.5 Hz          & 1.75 Hz           & 0.5--1.5 Hz     & Overall\\ \midrule
            15 cm  &        0/5( 0\%) & 1/5(20\%)       &       3/5(60\%)  &       3/5(60\%)  & \emph{5/5(100\%)}&\emph{5/5(100\%)}& \emph{5/5(100\%)} &       17/25(20\%)& 22/35(68\%)\\
            12 cm  & \emph{4/5(80\%)} &\emph{5/5(100\%)}& \emph{4/5(80\%)} & \emph{4/5(80\%)} & \emph{4/5(80\%)} & \emph{4/5(80\%)}& \emph{5/5(100\%)} &\emph{21/25(84\%)}&\emph{30/35(86\%)}\\ \midrule
            Overall&       4/10(40\%) &\emph{6/10(60\%)}& 7/10(70\%)       &       7/10(70\%) & \emph{9/10(90\%)}&\emph{9/10(90\%)}&\emph{10/10(100\%)}&       38/50(76\%)& 52/70(74\%)\\ \bottomrule
        \end{tabular}}
        }
        \label{wpaug3}
      \end{table}

  \subsection{Discussions}
    From the results obtained in the experiments, real-world data augmentation contributed to higher velocity-wise accuracy compared with simple changes in speed and improved the success rate with more repetitions of real-world playbacks.
    As the variation in speed in human demonstrations was less than 10\% for each task, real-world data augmentation also succeeded in extrapolating the velocity with which NN models can struggle. 
    Although the results did not follow the label perfectly, more samples of diverse playbacks or combinations with self-supervised learning~\cite{CRANEX7params} can enable fine-tuning. 

    This method has some limitations. 
    First, as this method relies on simple teaching--playback, the naive use of this method can only contribute to an increased diversity of speed and not position.
    Second, as teaching--playback does not incorporate feedback from the environment, it cannot be applied to tasks that require continued feedback, such as carrying cups full of water without spilling.
    
    Our future work will include the incorporation of this method with simulation-based data augmentation through the recreation of force information with autoregression, towards data augmentation for contact-rich tasks with variable positions in addition to variable speed.

\section{Conclusion}
  Because the world is not necessarily linear, changes in speed can result in hard-to-simulate differences in real-world responses that conventional data augmentation can encounter.
  In particular, the use of force control in imitation learning has made a real-world dataset of the contact force necessary, which is difficult to obtain through simulations.

  In this paper, we propose real-world data augmentation for imitation learning at variable speeds through teaching--playback, and we verified its effect on imitation learning.
  The proposed method, combined with bilateral control-based imitation learning, enables contact-rich manipulation at variable speeds from only two fixed-speed human demonstrations.
  Experimental results showed that the proposed method contributed to accuracy and stability along with command input by obtaining environmental reactions at a variety of speeds and higher success rates by multiplying real-world playbacks.
  
  We expect that a combination of this method and fine-tuning through self-supervised learning will enable the generation of accurate variable-speed motions in imitation learning.
  We also expect that this will enable generative data augmentation for both variable positions and variable speeds through a combination of the obtained data and simulation-based data augmentation.

\section*{Disclosure statement}

  The authors declare no potential conflicts of interest.

\section*{Funding}

  This work was supported by JSPS KAKENHI Grant Number 24K00905 and JST-ALCA-Next Japan, Grant Number JPMJAN24F1. This study was based on the results obtained from the JPNP20004 project subsidized by the New Energy and Industrial Technology Development Organization (NEDO).

\section*{Notes on contributor(s)}

  Nozomu Masuya received the B.E. degree in engineering systems from the University of Tsukuba, Japan, in 2023. He is currently working on an M.E. degree in intelligent and mechanical interaction systems at the Graduate School of Science and Technology at the University of Tsukuba. His research interests include motion control, robotics, artificial intelligence, and machine learning. 

  Hiroshi Sato received the B.E. degree from Shibaura Institute of Technology, Tokyo, Japan, in 2023. He is currently working on an M.E. degree in intelligent and mechanical interaction systems at the Graduate School of Science and Technology at the University of Tsukuba, Japan. His research interests include motion control, robotics, and machine learning.

  Koki Yamane received the B.E. degree in engineering systems and M.E. degree in intelligent and mechanical interaction systems from the University of Tsukuba, Japan, in 2022 and 2024, respectively. He is currently pursuing a Ph.D. in intelligent and mechanical interaction systems at the University of Tsukuba, Japan. His research interests include motion control, robotics, image processing, and machine learning.

  Takuya Kusume received the B.E. degree from the National Institution for Academic Degree and University Evaluation, Japan, in 2022, and M.E. degree in intelligent and mechanical interaction systems from the University of Tsukuba, Japan, in 2024.  

  Sho Sakaino received the B.E. degree in system design engineering and the M.E. and Ph.D. degrees in integrated design engineering from Keio University, Yokohama, Japan, in 2006, 2008, and 2011, respectively. He was an assistant professor at Saitama University between 2011 and 2019. Since 2019, he has been an associate professor at the University of Tsukuba. His research interests include mechatronics, motion control, robotics, and haptics. He received the IEEE IES Best Conference Paper Award in 2022. He also received the IEEJ Industry Application Society Distinguished Transaction Paper Award in 2011 and 2020 and the RSJ Advanced Robotics Excellent Paper Award in 2020.

  Toshiaki Tsuji received the B.E. degree in system design engineering and the M.E. and Ph. D degrees in integrated design engineering from Keio University, Yokohama, Japan, in 2001, 2003, and 2006, respectively. He was a research associate in the Department of Mechanical Engineering, Tokyo University of Science, from 2006 to 2007. He is currently an associate professor in the Department of Electrical and Electronic Systems, Saitama University. His research interests include motion control, haptics, and rehabilitation robots. He received the FANUC FA and Robot Foundation Original Paper Award in 2007 and 2008, respectively. He also received the RSJ Advanced Robotics Excellent Paper Award and IEEJ Industry Application Society Distinguished Transaction Paper Award in 2020.

\bibliographystyle{tfnlm}
\bibliography{masuya.bib}

\bigskip

\end{document}